\documentclass[sigconf]{acmart}

\usepackage{multirow}
\usepackage{subcaption}
\usepackage{algorithm}
\usepackage{algpseudocode}
\usepackage{alltt} 
\usepackage{enumitem}
\usepackage[most]{tcolorbox}
\algrenewcommand\algorithmiccomment[1]{\hfill$\triangleright$ #1}

\definecolor{promptGray}{RGB}{245,247,250}
\definecolor{promptBorder}{RGB}{86,110,155}
\newtcolorbox{PromptBox}[2][promptGray]{
  enhanced jigsaw,
  breakable,
  sharp corners,
  boxrule=0.7pt,
  colback=#1,
  colframe=promptBorder,
  title={#2},
  fonttitle=\bfseries,
  fontupper=\small,
  left=5pt,
  right=5pt,
  top=5pt,
  bottom=5pt,
  boxsep=3pt,
  before skip=6pt,
  after skip=6pt
}

\AtBeginDocument{%
  }

\copyrightyear{2026}
\acmYear{2026}
\setcopyright{cc}
\setcctype{by}
\acmConference[KDD '26]{Proceedings of the 32nd ACM SIGKDD Conference on Knowledge Discovery and Data Mining V.2}{August 09--13, 2026}{Jeju Island, Republic of Korea}
\acmBooktitle{Proceedings of the 32nd ACM SIGKDD Conference on Knowledge Discovery and Data Mining V.2 (KDD '26), August 09--13, 2026, Jeju Island, Republic of Korea}
\acmDOI{10.1145/3770855.3817982}
\acmISBN{979-8-4007-2259-2/2026/08}

\begin{document}

\title{ProfiliTable: Profiling-Driven Tabular Data Processing via Agentic Workflows}

\author{Wei Liu}
\authornote{The first three authors contributed equally to this research.}
\affiliation{%
  \institution{School of CS \& Key Lab of High Confidence Software Technologies (MOE), Peking University}
  \city{Beijing}
  \country{China}
}
\email{eularioal@stu.pku.edu.cn}

\author{Yang Gu}
\authornotemark[1]
\affiliation{%
  \institution{Academy for Advanced Interdisciplinary Studies, Peking University}
  \city{Beijing}
  \country{China}
}
\email{gu_yang@pku.edu.cn}


\author{Xi Yan}
\authornotemark[1]
\affiliation{%
  \institution{Institute of Computing Technology, Chinese Academy of Sciences}
  \city{Beijing}
  \country{China}
}
\email{yanxi25s@ict.ac.cn}

\author{Zihan Nan}
\affiliation{%
  \institution{School of Software \& Microelectronics, Peking University}
  \city{Beijing}
  \country{China}
}
\email{zhnan25@stu.pku.edu.cn}

\author{Beicheng Xu}
\affiliation{%
  \institution{School of CS \& Key Lab of High Confidence Software Technologies (MOE), Peking University}
  \city{Beijing}
  \country{China}
}
\email{beichengxu@stu.pku.edu.cn}

\author{Keyao Ding}
\affiliation{%
  \institution{School of CS \& Key Lab of High Confidence Software Technologies (MOE), Peking University}
  \city{Beijing}
  \country{China}
}
\email{maodeshi@stu.pku.edu.cn}

\author{Bin Cui}
\authornote{Corresponding authors.}
\affiliation{%
  \institution{School of CS \& Beijing Key Laboratory of Software and Hardware Cooperative Artificial Intelligence Systems, Peking University}
  \city{Beijing}
  \country{China}
}
\email{bin.cui@pku.edu.cn}

\author{Wentao Zhang}
\authornotemark[2]
\affiliation{%
  \institution{Center for Machine Learning Research, Peking University}
  \city{Beijing}
\country{China}
}
\affiliation{
\institution{Zhongguancun Academy}
\city{Beijing}
\country{China}
}
\email{wentao.zhang@pku.edu.cn}



\renewcommand{\shortauthors}{Wei Liu et al.}

\begin{abstract}

Table processing—including cleaning, transformation, augmentation, and matching—is a foundational yet error-prone stage in real-world data pipelines. While recent LLM-based approaches show promise for automating such tasks, they often struggle in practice due to ambiguous instructions, complex task structures, and the lack of structured feedback, resulting in syntactically correct but semantically flawed code. To address these challenges, we propose ProfiliTable, an autonomous multi-agent framework centered on dynamic profiling, which constructs and iteratively refines a unified execution context through interactive exploration, knowledge-augmented synthesis, and feedback-driven refinement. ProfiliTable integrates (i) a Profiler that performs ReAct-style data exploration to build semantic understanding, (ii) a Generator that retrieves curated operators to synthesize task-aware code, and (iii) a joint Evaluator–Summarizer module that injects execution scores and diagnostic insights to enable closed-loop refinement. Extensive experiments on a diverse benchmark covering 18 tabular task types demonstrate that ProfiliTable consistently outperforms strong baselines, particularly in complex multi-step scenarios. These results highlight the critical role of dynamic profiling in reliably translating ambiguous user intents into robust and governance-compliant table transformations. 

\end{abstract}




\maketitle

\newcommand\kddavailabilityurl{https://doi.org/xxxx}
\ifdefempty{\kddavailabilityurl}{}{
\begingroup\small\noindent\raggedright\textbf{Resource Availability:}\\
Extended version with full appendices: \url{https://arxiv.org/abs/2605.12376}.

The source code is available at \url{https://github.com/PKU-DAIR/ProfiliTable}.

\endgroup
}

\section{Introduction}
Tabular data constitutes the backbone of modern data-driven decision making and underpins a wide range of machine learning applications, especially in high-stakes domains such as finance \cite{apply:finance}, healthcare \cite{apply:healthCare}, and government \cite{apply:Gov}. However, real-world tabular data is rarely analysis-ready: it is often noisy, incomplete, and semantically inconsistent, requiring extensive processing before reliable modeling can begin \cite{apply:raw-data-dirty}. Moreover, the diversity of table structures and domain semantics precludes the existence of a single, fixed algorithm that can handle all data wrangling scenarios \cite{apply:no-one-size-all-model}. As a result, rule-based table processing tools such as OpenRefine \cite{openrefine} demand substantial manual configuration and expert intervention, making tabular data processing one of the most labor-intensive and error-prone stages in the data science pipeline \cite{polyzotis2017data, SambasivanCHI, Ridzuan_2019}.


These limitations have motivated growing interest in more flexible and automated solutions that can adapt to diverse tabular structures and task requirements with minimal manual intervention. The advent of large language models (LLMs) has opened new avenues for automating data-centric tasks \cite{LLM-data-centric1}. Recent works have demonstrated LLMs’ ability to generate executable code for table manipulation from natural language, enabling data science agents that produce pandas or SQL scripts \cite{trirat2025automlagent, liu2025mlMasteraiforaiintegrationexploration, MLCopilot}. However, existing approaches fall short in delivering robust, general-purpose table curation for three key reasons: First, many methods are narrowly scoped, focusing on isolated operations such as imputation \cite{srinivasan2025doespromptdesignimpact, Data-Impute2, Data-Impute3}, error correction \cite{bendinelli2025exploring, IterClean, LLMClean}, or schema matching \cite{fu2025incontextclusteringbasedentityresolution}, and thus cannot handle composite workflows \cite{AutoDCWorkflow, zhu2025relational}. Second, broader frameworks often prioritize other data science tasks—such as automated machine learning (AutoML) \cite{automlgpt, dsMentor} or table-based question answering \cite{Table-LLM-Specialist, Table-Meets-LLM}—over the full spectrum of table processing. Third, even those systems explicitly targeting table processing either rely on generic multi-agent scaffolds, which lack table-specific data understanding \cite{MetaGPT, ChatDev}, or employ hand-crafted pipelines that do not emphasize adaptive, feedback-driven profiling \cite{govbench, CleanAgent}.

These limitations persist in current tools. For example, CleanAgent performs column-type annotation but does not actively sample or inspect the actual cell values during code generation \cite{CleanAgent}. As a result, when given an ambiguous instruction such as “standardize the currency column”, it has no knowledge of the concrete values present in the column and therefore cannot determine how to construct an appropriate mapping table for standardization.

In contrast, a rule-based system like DataGovAgent typically samples only a few rows per column to get basic properties \cite{govbench}. However, this limited inspection often fails to capture the full range of values needed to interpret ambiguous instructions—especially when the relevant column cannot be identified without examining its actual content. To handle such cases reliably, it would need to compute full value statistics such as unique values for that column. But since the target column is unknown in advance, the only safe option is to compute unique values for all columns. This leads to significant redundancy, especially in wide tables, and floods the prompt with irrelevant information. The root issue is that static, rule-driven profiling cannot decide adaptively which column to explore based on the instruction, and therefore must over-provision to avoid failure.

These shortcomings highlight why table processing demands a flexible and hypothesis-driven profiling mechanism that explores only what is necessary and does so at the right time. Therefore, we propose \textbf{ProfiliTable}, whose core insight is that robust table processing depends on an \textbf{active, iterative notion of profiling}—not as passive metadata consumption, but as a dynamic process of semantic construction that evolves through interaction, feedback, and knowledge integration. Our profiling framework operates through three synergistic mechanisms:  (1) \textit{Interactive exploration}: The agent actively interrogates tables via a ReAct loop \cite{ReAct}, executing lightweight actions (e.g., sampling) to formulate and test hypotheses about data quality and structure, going beyond the \textbf{fixed, one-size-fits-all rules used in prior works} \cite{govbench};  (2) \textit{Knowledge augmented synthesis}: For complex operations, the agent decomposes tasks into subgoals and retrieves relevant, pre-validated operator templates from a curated library via Retrie\-val Augmented Generation (RAG) \cite{rag}, ensuring that generated code is based on reliable, domain-specific primitives. (3) \textit{Feed\-back-driven refinement}: After each code generation attempt, the Evaluator computes a task-specific score using ground-truth-aligned logic. A dedicated Summarizer agent then performs a ReAct-style loop over the processed table—formulating verification hypotheses, executing lightweight validation actions, and observing outcomes—to assess alignment with the original intent. The resulting insight is fed to the Profiler and Generator, guiding more accurate profiling and code generation.

To evaluate table processing tasks, existing benchmarks for LLM-based table processing often misalign with real-world needs: they either focus on end-to-end analytics or isolated code correctness \cite{ml-bench, datascibench}, or lack support for multi-step table transformation workflows and semantics required in real production systems \cite{tablebench, treb, ds-1000}. DataGovBench \cite{govbench} introduces a hierarchical evaluation with rigorous metrics, but underutilizes classical processing operations. We extend it into a more comprehensive benchmark with fine-grained categories—covering cleaning, transformation, augmentation, and matching—to better reflect real-world table curation challenges. For performance, ProfiliTable consistently achieves state-of-the-art performance across both single-step and multi-step tasks under GPT-4o and GPT-5.2, outperforming all baselines in correctness, completeness, and execution reliability. Notably, it achieves a 100\% task-wise run\-nable rate—a key requirement for production deployment.

Our main contributions can be summarized as follows:  
\begin{itemize}[left=0pt]
    
    \item We propose ProfiliTable, an autonomous multi-agent framework that unifies interactive exploration, feedback-driven refinement, and knowledge-augmented synthesis under a \textbf{dynamic profiling paradigm} to overcome the brittleness of one-shot LLM code generation.  
    \item We construct a benchmark comprising 18 distinct types of table processing tasks—spanning cleaning, transformation, augmentation, and matching.
    \item Extensive experiments demonstrate ProfiliTable’s strong performance across our benchmark, achieving \textbf{state-of-the-art} results in both single-step and complex multi-step settings, with consistent gains in correctness, completeness, and execution reliability over strong baselines.
\end{itemize}

\section{Related Work}

\textbf{Table-Oriented Agents.}
Large language models have evolved into autonomous agents specialized for tabular data tasks. Recent systems focus explicitly on understanding, cleaning, and transforming tables: Data Interpreter~\cite{DataInterpreter} enables end-to-end analysis of structured data through code generation; DataGovAgent~\cite{govbench} ensures data governance via contract-guided planning; and CleanAgent~\cite{CleanAgent} standardizes messy tables through declarative APIs. Complementing these efforts, Li et al.~\cite{LLM-TabAD} formulate anomaly detection as a zero-shot batch-level reasoning task, showing that LLMs can identify low-density regions in numerical tables without model retraining. Similarly, SheetMind~\cite{sheetmind} facilitates robust spreadsheet automation via a multi-agent architecture that translates natural language into grammar-constrained executable commands, ensuring alignment with user intent through reflective validation.
These approaches collectively demonstrate a shift toward agents that treat tables as first-class citizens—reasoning over their schema, semantics, quality, and statistical structure.

\textbf{Multi-Agent LLM Frameworks.}
Beyond single-agent architectures, recent works explore collaborative multi-agent systems where specialized agents interact to solve complex tasks. Minimum-function agents—such as those in CAMEL \cite{CAMEL} and ChatDev \cite{ChatDev}—assign narrow roles (e.g., coder, tester) to reduce cognitive load, while client-server frameworks employ a central controller to plan workflows and delegate subtasks to client agents, improving modularity and robustness \cite{automlgpt, FinRobot, MetaGPT, xu2025llm}. More adaptive approaches include dynamic agent systems that create or reconfigure agents at runtime: hierarchical generation spawns child agents from a parent controller for structured decomposition \cite{Self-Organized-Agents}, while iterative feedback-driven methods (e.g., EvoMAC \cite{EvoMAC}) refine agent behavior through cross-agent error signals. Although dynamic designs offer flexibility, they often sacrifice stability and reproducibility. Our work uses role-specialized agents such as Interpreter, Profiler, Generator, and Summarizer. These agents operate within a closed-loop, feedback-driven pipeline that balances structured collaboration and adaptive refinement for table processing.

\section{ProfiliTable}
\subsection{Problem Formulation}
We study the problem of automated table processing: given a \textbf{natural language instruction} \( i \in I \) (e.g., ``impute missing values'') and raw input tables \( \mathcal{D}_{\text{raw}} \), the goal is to produce a \textbf{processed table} \( T \) that correctly fulfills the intent while satisfying syntactic and semantic validity as well as domain-specific constraints.

We aim to build an \textbf{agentic workflow} that, given a natural language instruction and a raw table, automatically produces a correctly processed output table. The workflow should maximize performance across real-world table processing tasks by generating outputs that closely match expert-validated ground-truth results.

In practice, real-world tables are often \textbf{too large to be loaded} entirely into memory. The workflow therefore does not receive direct access to \( \mathcal{D}_{\text{raw}} \); instead, it is only provided with file paths and must interact with the data through a controlled programming interface—such as sampling rows.

Consequently, the workflow generates executable code \( c \) such that \( T = \texttt{exec}(c, \mathcal{D}_{\text{raw}}) \), rather than producing the output table \( T \) directly. However, due to the ambiguity of natural language instructions and the complexity of tabular semantics, a one-shot code generation strategy often fails to yield correct or robust results. To address this, we model the workflow as an \textbf{iterative refinement system} that engages in closed-loop interaction with the execution environment, progressively improving its solution through feedback.

Let \( t = 1, 2, \dots, T_{\max} \) denote the iteration index. At each step \( t \), the workflow produces a candidate program \( c^{(t)} \) based on the original instruction \( i \), the raw data \( \mathcal{D}_{\text{raw}} \), local memory \( \mathcal{M}^{(t)} \), and a unified \textbf{dynamic profiling context} \( \mathcal{P}^{(t)} \), defined as:
\begin{equation}
\mathcal{P}^{(t)} = \Big( \underbrace{p^{(t)}}_{\text{current profiling}},\ 
\underbrace{\mathcal{R}^{(t)}}_{\text{retrieved operators}},\ 
\underbrace{\mathcal{F}^{(t)}}_{\text{feedback history}} \Big),
\end{equation}
where:
\begin{itemize}[left=0pt]
    \item \( p^{(t)} \) is the active profiling summary generated by the Profiler through ReAct-style exploration of \( \mathcal{D}_{\text{raw}} \);
    \item \( \mathcal{R}^{(t)} \) denotes operator templates retrieved from a curated knowledge base via RAG;
    \item \( \mathcal{F}^{(t)} = \{ f^{(1)}, \dots, f^{(t-1)} \} \) is the feedback history from previous iterations, with each feedback signal
    \begin{equation}
    f^{(m)} = \big( s^{(m)},\, e^{(m)},\, \tau^{(m)},\, p^{(m)} \big),
    \end{equation}
    comprising: (i) execution score \( s^{(m)} = \text{Eval}\big( \texttt{exec}(c^{(m)}, \mathcal{D}_{\text{raw}}) \big) \),  
    (ii) error trace \( e^{(m)} \) (if any),  
    (iii) diagnostic insight \( \tau^{(m)} \) from the Summarizer’s ReAct-based validation, and  
    (iv) the profiling summary \( p^{(m)} \) used at step \( m \).
\end{itemize}
The dynamic profiling context $\mathcal{P}^{(t)}$ evolves iteratively, as illustrated in Figure~\ref{fig:case-study}, becoming increasingly informative and better aligned with the user's task-specific intent.

The code generation policy is thus \textbf{profiling-aware}:
\begin{equation}
\label{eqa:ct}
c^{(t)} = g\big( i,\, \mathcal{D}_{\text{raw}},\, \mathcal{P}^{(t)},\, \mathcal{M}^{(t)} \big),
\end{equation}
where the function \( g \) is implemented by a multi-agent collaboration pipeline. The overall self-improving workflow is depicted in Figure \ref{fig:architecture}, highlighting the closed-loop refinement loop and the roles of each agent across iterations \( t = 1, 2, \dots, T_{\max} \).

\subsection{Autonomous Multi-Agent Framework}

\begin{algorithm}[t]
\caption{ProfiliTable: Dynamic Profiling Workflow}
\label{alg:profilitable}
\small
\setlength{\itemsep}{0pt}
\begin{algorithmic}[1]
\State Parse instruction $i$ with Interpreter to get task complexity.
\State Initialize $\mathcal{P}^{(0)} \gets (\emptyset, \emptyset, \emptyset)$, $\mathcal{F}^{(1)} \gets \emptyset$, $\mathcal{M}^{(1)} \gets \emptyset$, best\_score $\gets -\infty$, best\_code $\gets \emptyset$.
\For{$t = 1$ to $T_{\max}$}
    \State $p^{(t)} \gets$ Profiler($\mathcal{D}_{\text{raw}}$, $\mathcal{P}^{(t-1)}$) \Comment{explore raw table (ReAct)}
    \If{task is multi-step}
        \State $\{i_m\} \gets$ Decomposer($i$) \Comment{decompose task}
        \State $\mathcal{R}^{(t)} \gets \bigcup_m \text{top-}k\text{ operators for } i_m$
    \Else
        \State $\mathcal{R}^{(t)} \gets \text{top-}k\text{ operators for } i$ \Comment{retrieve operators}
    \EndIf
    \State $\mathcal{P}^{(t)} \gets (p^{(t)}, \mathcal{R}^{(t)}, \mathcal{F}^{(t)})$ \Comment{build profiling context}
    \State $c^{(t)} \gets$ Generator($i$, $\mathcal{D}_{\text{raw}}$, $\mathcal{P}^{(t)}$, $\mathcal{M}^{(t)}$) \Comment{generate code}
    \State $\mathcal{M}^{(t+1)} \gets$ UpdateMemory($\mathcal{M}^{(t)}$, $c^{(t)}$)
    \State $(T^{(t)}, e^{(t)}) \gets \texttt{exec}(c^{(t)}, \mathcal{D}_{\text{raw}})$ \Comment{execute code and collect errors}
    \State $s^{(t)} \gets$ Evaluator($T^{(t)}$, $i$, $e^{(t)}$) \Comment{set $s^{(t)}=0$ if execution fails}
    \State $\tau^{(t)} \gets$ Summarizer($T^{(t)}$, $i$, $s^{(t)}$, $e^{(t)}$) \Comment{validate result (ReAct)}
    \State $f^{(t)} \gets (s^{(t)}, e^{(t)}, \tau^{(t)}, p^{(t)})$
    \State $\mathcal{F}^{(t+1)} \gets \mathcal{F}^{(t)} \cup \{ f^{(t)} \}$ \Comment{update feedback}
    \If{$s^{(t)} >$ best\_score}
        \State best\_score $\gets s^{(t)}$, best\_code $\gets c^{(t)}$
    \EndIf
    \If{$s^{(t)} \geq \theta$} \Comment{score surpasses threshold}
        \State \Return $T^{(t)}$ \Comment{return current output}
    \EndIf
\EndFor
\State $T^* \gets \texttt{exec}(\text{best\_code}, \mathcal{D}_{\text{raw}})$ \Comment{return best}
\State \Return $T^*$
\end{algorithmic}
\end{algorithm}

\begin{figure}[t]
    \centering
    \includegraphics[width=1\linewidth]{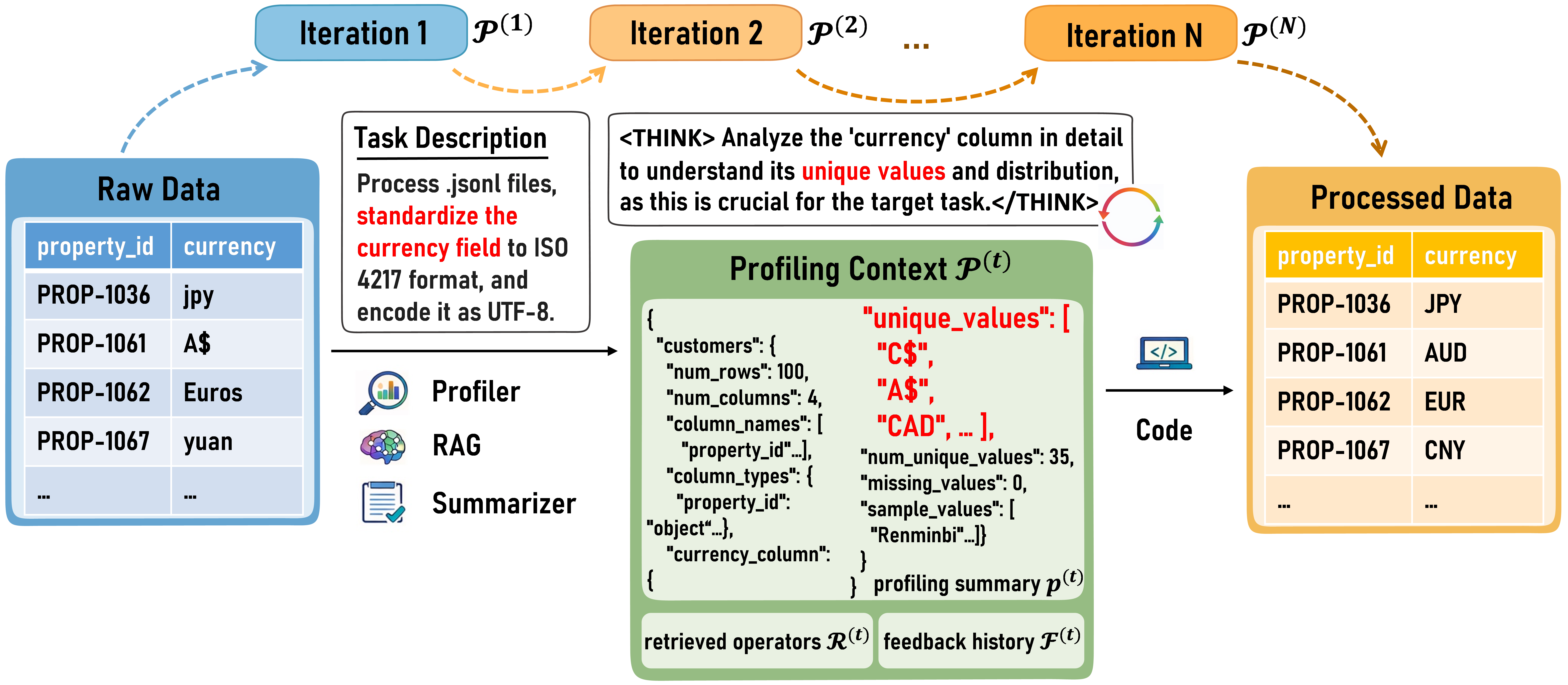}
    \caption{Profiling reveals ambiguous instructions and recovers concrete currency symbols for accurate ISO 4217 mapping.}
    \label{fig:case-study}
\end{figure}

\begin{figure*}[t]
    \centering
    
    \includegraphics[width=0.92\textwidth]{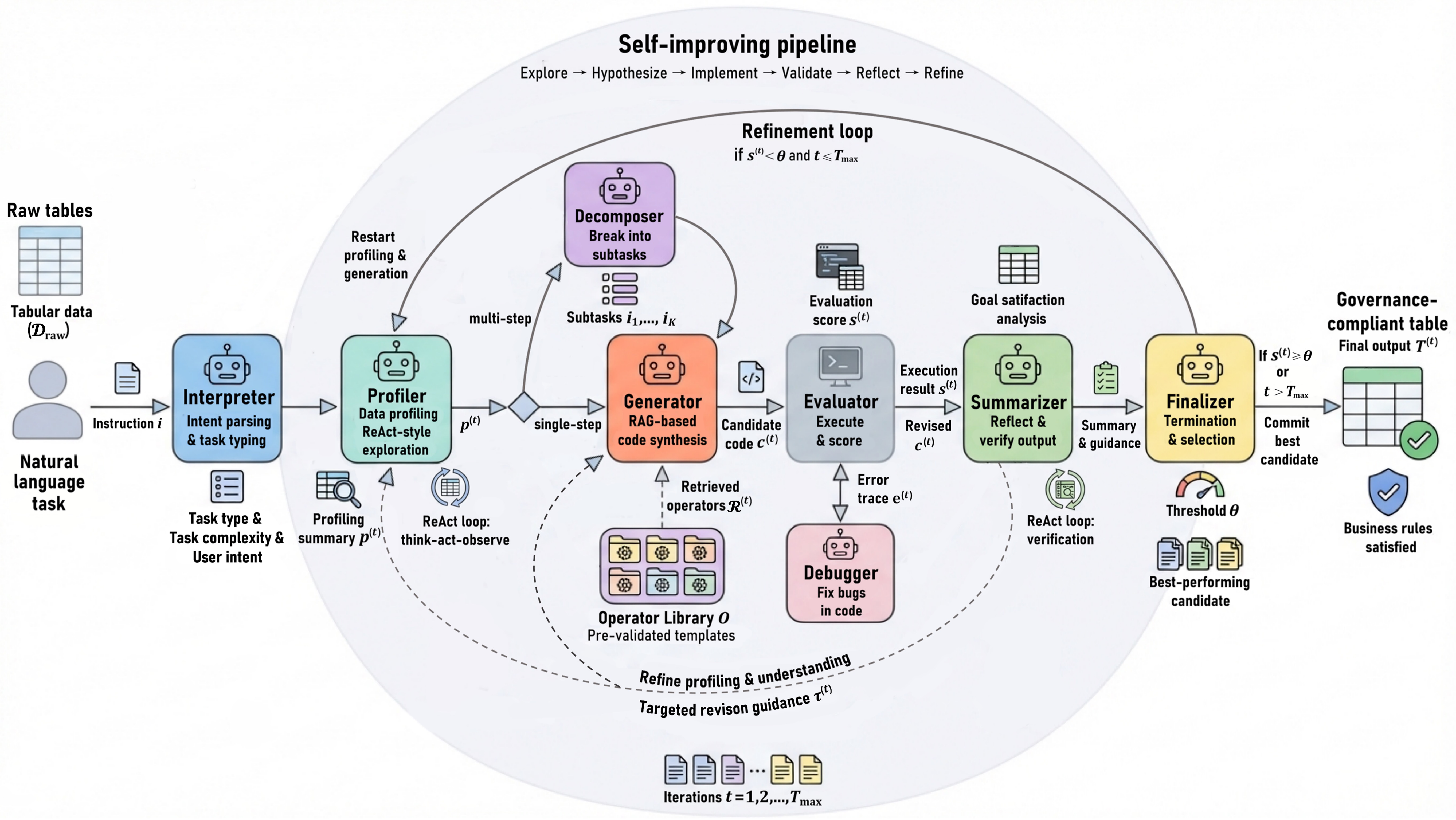}
    \caption{The autonomous workflow of ProfiliTable: a self-improving, closed-loop pipeline centered around a dynamic \textbf{profiling context}. The Interpreter parses user intent; the Profiler conducts ReAct-style data exploration to initialize the context; the Decomposer decomposes multi-step tasks; the Generator synthesizes code by grounding in retrieved operators and the current profiling summary; the Evaluator and Summarizer jointly enrich the context with execution feedback and diagnostic insights; and the Finalizer selects the best candidate upon convergence—enabling robust and governance-compliant table processing.}
    \label{fig:architecture}
\end{figure*}
As illustrated in Figure \ref{fig:architecture} and algorithm \ref{alg:profilitable}, our table processing workflow implements a closed-loop, feedback-driven framework through several specialized LLM-powered agents: Interpreter, Profiler, Decomposer, Generator, Evaluator, Summarizer and Finalizer. This architecture enables robust, interpretable, and iterative transformation of raw tables into governance-compliant outputs.

\textbf{Interpreter Agent.} The Interpreter is the first component in the workflow. It clarifies the user's intent and identifies the task type and complexity—determining whether the request is single-step or multi-step. This decision directly controls operator retrieval: for single-step tasks, it selects operators that match the task type; for multi-step tasks, it forwards the request to the Decomposer. The Decomposer breaks the task into subtasks and retrieves matching operators for each subtask. By accurately recognizing user intent, the Interpreter ensures that downstream modules act with the right focus and level of detail.

\textbf{Profiler Agent.} The Profiler acts as the semantic grounding module, actively interrogating input tables to build a contextual understanding of their structure and semantics. Given a natural language instruction \(i \in I\) and raw tables \(\mathcal{D}_{\text{raw}}\), it runs a \textbf{ReAct-style exploration loop} \cite{ReAct}. As illustrated in \autoref{fig:case-study}, it recognizes that resolving ambiguities—such as ``standardize the currency field''—requires inspecting actual values (e.g., unique values) rather than relying on schema metadata alone. The output is a \textbf{profiling summary} \(p^{(t)}\), which encodes distributional properties, data quality issues, and key observations, and is passed to downstream agents to ensure data-aware code generation and bridge linguistic intent with tabular reality.

To illustrate this mechanism, consider the case shown in Figure \ref{fig:case-study}: for complex instructions like “standardize the currency field to ISO 4217 format”, the Profiler first identifies the target column via semantic matching. Instead of relying on static sampling, it proactively queries all unique values to establish a comprehensive mapping. Leveraging LLM-based commonsense reasoning, it infers standard entities and verifies ISO 4217 compliance for each value. The resulting task-driven profiling summary, $p^{(t)}$, encodes deep semantic insights (e.g., mapping “yuan” and “RMB” to “CNY”) alongside basic statistics. By avoiding the information loss of static sampling methods~\cite{govbench}, this active exploration mechanism grounds downstream code generation in complete semantic context and improves both intent alignment and execution robustness.

\textbf{Decomposer Agent.} The Decomposer is triggered only for tasks that the Interpreter classifies as multi-step. It decomposes a complex instruction into an ordered sequence of atomic subtasks, each formalized as a (task type, description) tuple. This structured decomposition enables targeted operator retrieval for each subtask by aligning the task type and semantic description with entries in the operator library. Moreover, it provides downstream agents with granular, well-scoped objectives, enabling focused exploration, precise code generation, and step-wise validation. By decomposing composite requests into executable units, the Decomposer ensures reliable and coherent multi-step workflows.

\textbf{Generator Agent.} The Generator translates the user’s intent \(i\) and the unified \textbf{dynamic profiling context} \(\mathcal{P}^{(t)}\) into an executable program \(c^{(t)}\). It employs a task-aware retrieval-augmented strategy that adapts to the structural complexity of \(i\).

Let \(\mathcal{O} = \{o_1, \dots, o_M\}\) denote a curated library of pre-validated operator templates, each associated with a natural language description \(d(o_i)\). Given a similarity function \(\text{sim}(\cdot, \cdot)\) (cosine similarity over embedding vectors), retrieval proceeds as follows:

\begin{itemize}[left=0pt]
    \item For \textbf{single-step tasks}, the Generator treats \(i\) as an atomic query and retrieves operators satisfying:
    \begin{equation}
    \mathcal{R}_{\text{single}} = \left\{ o_i \in \mathcal{O} \,\middle|\, \text{sim}\big(d(o_i), i\big) \geq \theta_{\text{sim}} \right\},
    \end{equation}
    then selects the top-\(k\) most relevant ones:
    \begin{equation}
    \mathcal{R}^{(t)} = \text{top-}k\big(\mathcal{R}_{\text{single}}\big).
    \end{equation}
    \item For \textbf{multi-step tasks}, the Decomposer first decomposes \(i\) into subtasks \(\{i_1, \dots, i_K\}\). For each \(i_m\), it retrieves:
    \begin{equation}
    \mathcal{R}_m = \text{top-}k\left( \left\{ o_i \in \mathcal{O} \,\middle|\, \text{sim}\big(d(o_i), i_m\big) \geq \theta_{\text{sim}} \right\} \right),
    \end{equation}
    and forms the overall retrieval set as:
    \begin{equation}
    \mathcal{R}^{(t)} = \bigcup_{m=1}^K \mathcal{R}_m. 
    \end{equation}
\end{itemize}

The retrieved operators in \(\mathcal{R}^{(t)}\) are injected as in-context exemplars. The Generator then synthesizes the candidate program, as shown in Equation~(\ref{eqa:ct}).


\textbf{Evaluator Agent.} The Evaluator provides rigorous, task-specific validation by executing \(c^{(t)}\) in a sandboxed environment and computing a feedback signal:
\begin{equation}
s^{(t)} = \text{Eval}\big( \texttt{exec}(c^{(t)}, \mathcal{D}_{\text{raw}}) \big),
\end{equation}
where \(s^{(t)} \in [0,1]\) is the evaluation score. Critically, \(\text{Eval}(\cdot)\) is a \textbf{task-customized, non-LLM-based script} that compares the output against ground-truth (GT) expectations using precise logic (e.g., F1 score). To ensure strict isolation and prevent data leakage, GT access is confined exclusively to this evaluator module. The rest of the framework—including the Profiler and Generator—never accesses actual GT tables or evaluation logic; they only receive the scalar score \(s^{(t)}\) as feedback. This design ensures that ``runnable'' does not imply ``correct''—only solutions that satisfy business objectives receive high scores—while maintaining a realistic deployment scenario where GT is invisible to the reasoning pipeline.

\textbf{Summarizer Agent.} The Summarizer operates in a ReAct-style loop analogous to the Profiler, but with a distinct objective: rather than exploring raw data, it interacts directly with the processed table \(T^{(t)} = \texttt{exec}(c^{(t)}, \mathcal{D}_{\text{raw}})\) to assess whether the current output satisfies the original task objective \(i\). At each turn, the Summarizer formulates verification hypotheses (e.g., ``Has all missing `income' been imputed?''), executes lightweight validation actions over \(T^{(t)}\), and observes outcomes to determine task completion status. This interaction yields a structured assessment of alignment between the current output and the ground-truth expectations encoded in \(i\). Based on this assessment and the feedback history \(\mathcal{F}^{(t)}\), the Summarizer synthesizes a concise insight that explicitly identifies any unfulfilled subgoals and provides targeted revision guidance. This insight is then propagated back to both the Profiler and Generator, enabling the Profiler to refine its data understanding in subsequent iterations and guiding the Generator to adjust its code synthesis strategy. Through this reflection, the workflow maintains a coherent reasoning chain from intent to validation, grounding each refinement in both current output observations and past attempts.

\textbf{Finalizer Agent.} The Finalizer acts as the termination and selection module governing convergence of the iterative refinement loop. At each invocation, it evaluates the current state against two criteria: (i) whether the evaluation score \(s^{(t)} = \text{Eval}(T^{(t)})\) meets or exceeds a pre-specified success threshold \(\theta\), and (ii) whether the number of generation attempts has reached the maximum allowed retries \(T_{\max}\). If \(s^{(t)} \geq \theta\), the Finalizer halts the workflow and commits \(T^{(t)}\) as the solution. If \(s^{(t)} < \theta\) but retries remain, it returns control to the Profiler for another refinement cycle. Critically, if the retry limit is reached without achieving \(\theta\), the Finalizer does not return the latest attempt; instead, it selects the \textbf{best-performing candidate} across all iterations—i.e., the code \(c^{(t^*)}\) and corresponding table \(T^{(t^*)}\) with the highest score \(s^{(t^*)} = \max_{i \leq t} s^{(i)}\). This ensures that even in failure-to-converge cases, the workflow delivers the most accurate result observed during execution.

Together, these agents form a \textbf{self-improving pipeline} that mirrors human data wrangling: explore → hypothesize → implement → validate → reflect → refine. At the heart of this loop is a unified \textbf{dynamic profiling context}, which continuously integrates real-time data exploration, retrieved operator knowledge, and multi-round feedback into a coherent semantic representation. By grounding every generation and validation step in this evolving context, our workflow achieves high-fidelity table processing while maintaining full interpretability and auditability.
\section{Experiment}
\subsection{Experiment Settings}
\subsubsection{Benchmark}
Our experiments are conducted on a comprehensive benchmark of various table processing tasks, carefully designed to reflect the diverse challenges encountered in practical data curation workflows. Each task is specified by a natural language instruction, accompanied by one or more raw input files (in CSV or JSONL format), a ground-truth output file, and a dedicated evaluation script (\texttt{eval.py}) that computes a normalized score in \([0,1]\) by comparing the agent’s output against the expected result.  

The benchmark encompasses 18 fine-grained task categories organized into four core dimensions: Table Cleaning (error correction, data imputation), Table Transformation (formatting, standardization, normalization, mapping, aggregation, concatenation, splitting, pivoting, column/row swapping, filtering, grouping, sorting), Table Augmentation (row population, schema evolution), and Table Matching (schema alignment, entity resolution). Tasks are categorized as either single-step, resolvable by a single atomic operation, or multi-step, requiring the orchestration of multiple operations into a coherent pipeline. Comprising 90 single-step and 39 multi-step tasks, our benchmark extends DataGovBench by retaining existing table-related tasks and introducing 19 new ones for comprehensive coverage. The single-step tasks are distributed across the four dimensions as follows: Table Transformation (38 tasks), Table Cleaning (35), Table Matching (13), and Table Augmentation (4).

\subsubsection{Metrics}
We evaluate table processing agents using two complementary categories of metrics: \textbf{performance metrics}, which assess correctness, robustness, and solution quality, and \textbf{efficiency metrics}, which quantify computational cost and latency. All performance metrics are scaled by a factor of 100 (i.e., reported as percentages) for readability. The Average Task Score (ATS) computes the mean of normalized task-specific scores across all evaluated instances, where each score reflects the alignment between the generated output and ground-truth expectations. To assess perfect execution, we report the Task Success Rate (TSR), defined as the fraction of tasks achieving a full score of 1.0. Recognizing that many real-world tasks admit partial solutions, we further introduce the Partial Success Rate (PSR)—the proportion of tasks yielding a strictly positive score—as a measure of meaningful progress even in the absence of complete correctness. On the code reliability front, the Code Runnable Rate (CRR) quantifies the ratio of successfully executed code snippets to the total number of generated scripts, capturing syntactic validity and runtime stability. Complementing this, the Task-wise Runnable Rate (TRR) measures the fraction of tasks for which at least one attempt produces a runnable script, reflecting per-task resilience against generation failures. To summarize overall effectiveness, we further report the average metric score (Avg. Score), defined as the arithmetic mean of ATS, TSR, PSR, CRR, and TRR for each method. As for the \textbf{Efficiency metrics}, we report average tokens (Avg. Tokens) and average time (Avg. Time), which denote the average number of tokens consumed by agents and the average wall-clock time (in seconds) spent per task, respectively. Together, they provide a holistic view of agent performance, jointly reflecting computational cost, latency, and practical usability in complex table processing workflows. See Appendix~\ref{subsec:metric} for a summary.

\subsubsection{Baselines}
We compare ProfiliTable against several representative LLM-based agents that exemplify current trends in table-oriented tasks and multi-agent collaboration. DataGovAgent \cite{govbench} operates on the DataGovBench benchmark with contract-guided planning and retrieval-augmented generation to ensure reliable execution of data governance tasks. CleanAgent \cite{CleanAgent} focuses specifically on cleaning messy tables by applying declarative transformation rules derived from natural language instructions. On the multi-agent front, CAMEL \cite{CAMEL} and ChatDev \cite{ChatDev} simulate role-playing teams that coordinate via natural language dialogue to decompose and solve software development tasks, while MetaGPT \cite{MetaGPT} orchestrates agent collaboration through standardized workflows and shared memory to manage complex project pipelines. DeepAnalyze \cite{DeepAnalyze} supports exploratory data analysis via a domain-fine-tuned LLM, but lacks a multi-agent architecture with distinct roles; we include its results in the Appendix \ref{subsec: add-exp} for completeness. Together, these baselines reflect the spectrum of approaches—from single-agent reasoning to multi-agent teamwork—that underpin modern LLM-driven automation in data-centric domains. 

\subsubsection{Implementation Details}
All experiments are conducted on our benchmark, with each method evaluated using both GPT-4o and GPT-5.2. For baseline frameworks, we use their default inference configurations as reported in their original papers. For ProfiliTable, we employ its full pipeline. Its operator library is organized hierarchically by task category, including both major and sub-categories. For example, within the \textit{Error Detection and Correction} sub-category of Table Cleaning, we include operators such as \textit{mass\_edit\_values} for batch editing of table entries. The operators are primarily derived from classic table processing libraries, such as OpenRefine and pandas; we adapt relevant functionalities from these sources into Python scripts, manually review them, and then incorporate them into our library. We construct at least four operators for each sub-task category, resulting in 68 operators in total. During inference, at most 2 operator templates are retrieved via RAG with a similarity threshold of 0.5; the ReAct-style profiling loop runs for up to 7 steps; the overall feedback-driven refinement is repeated for at most 3 rounds with a score threshold of 0.8; and the debugging sub-routine is allowed up to 5 attempts per iteration. Individual tasks are allowed to run to completion, subject to a 30-minute time cap. To ensure fairness, we impose no restrictions on tool usage or internal mechanisms, allowing each framework to execute its original pipeline under the same execution budget. 

\subsection{Overall Performance}

\begin{table*}[ht]
\centering
\small
\caption{Performance comparison on single-step tasks. Best results are in \textbf{bold}, and second-best results are \underline{underlined}.}
\label{tab:op_results}
\begin{tabular}{lccccccccc}
\toprule
\textbf{Base Model} & \textbf{Framework} & \textbf{ATS$\uparrow$} & \textbf{TSR$\uparrow$} & \textbf{PSR$\uparrow$} & \textbf{CRR$\uparrow$} & \textbf{TRR$\uparrow$} & \textbf{Avg. Score$\uparrow$} & \textbf{Avg. Tokens$\downarrow$} & \textbf{Avg. Time (s)$\downarrow$} \\
\midrule

\multirow{7}{*}{GPT-4o} 
& MetaGPT      & \underline{56.21} & 44.44 & \underline{62.22} & \underline{57.66} & 75.56 & \underline{59.22} & \textbf{15,241} & 69.18 \\
& CAMEL        & 46.83 & 33.33 & 54.44 & 44.83 & 72.22 & 50.33 & 82,845 & 70.56 \\
& CleanAgent    & 37.03 & 30.00 & 46.67 & 57.14 & \underline{78.89} & 49.95 & \underline{18,043} & \textbf{17.93} \\
& ChatDev2.0   & 55.89 & 44.44 & 61.11 & 52.5 & 70.00 & 56.79 & 114,785 & 133.23 \\
& DataGovAgent     & 52.34 & \underline{45.56} & 56.67 & 54.46 & 67.78 & 55.36 & 28,501 & 51.56 \\

& \textbf{ProfiliTable (ours)} & \textbf{86.82} & \textbf{68.89} & \textbf{93.33} & \textbf{84.72} & \textbf{97.78} & \textbf{86.31} & 25,794 & \underline{51.44} \\
\midrule

\multirow{6}{*}{GPT-5.2} 
& MetaGPT      & 67.31 & 54.44 & 73.33 & 68.97 & \underline{87.78} & 70.37 & 74,985 & 115.05 \\
& CAMEL        & 61.14 & 47.78 & 68.89 & 73.27 & 82.22 & 66.66 & 36,828 & 72.74 \\
& CleanAgent    & 45.91 & 35.56 & 53.33 & 50.64 & 86.67 & 54.42 & 33,497 & \textbf{44.75} \\
& ChatDev2.0   & \underline{70.02} & 60.00 & \underline{75.56} & 72.82 & 83.33 & 72.35 & 120,849 & 170.08 \\
& DataGovAgent     & 67.94 & \underline{61.11} & 74.44 & \underline{80.43} & 82.22 & \underline{73.23} & \underline{29,965} & \underline{72.42} \\
& \textbf{ProfiliTable (ours)} & \textbf{89.66} & \textbf{73.33} & \textbf{94.44} & \textbf{87.88} & \textbf{97.78} & \textbf{88.62} & \textbf{24,907} & 130.26 \\
\bottomrule
\end{tabular}
\end{table*}

\begin{table*}[ht]
\centering
\small
\caption{Performance comparison on multi-step tasks. Best results are in \textbf{bold}, and second-best results are \underline{underlined}.}
\label{tab:dag_results}
\begin{tabular}{lccccccccc}
\toprule
\textbf{Base Model} & \textbf{Framework} & \textbf{ATS$\uparrow$} & \textbf{TSR$\uparrow$} & \textbf{PSR$\uparrow$} & \textbf{CRR$\uparrow$} & \textbf{TRR$\uparrow$} & \textbf{Avg. Score$\uparrow$} & \textbf{Avg. Tokens$\downarrow$} & \textbf{Avg. Time (s)$\downarrow$} \\
\midrule

\multirow{7}{*}{GPT-4o} 
& MetaGPT      & 49.53 & 24.32 & 59.46 & 45.61 & 70.27 & 49.84 & 30,038 & \underline{68.91} \\
& CAMEL        & 32.64 & 16.22 & 64.86 & 41.18 & 70.27 & 45.03 & 115,116 & 94.33 \\
& CleanAgent    & 46.03 & 21.62 & 62.16 & 44.07 & 70.27 & 48.83 & \textbf{21,562} & \textbf{20.59} \\
& ChatDev2.0   & \underline{57.26} & \underline{27.03} & \underline{75.68} & 65.96 & \underline{83.78} & \underline{61.94} & 133,688 & 143.77 \\
& DataGovAgent     & 47.98 & \underline{27.03} & 72.97 & \underline{72.50} & 78.38 & 59.77 & \underline{29,189} & 91.52 \\
& \textbf{ProfiliTable (ours)} & \textbf{80.19} & \textbf{45.95} & \textbf{94.59} & \textbf{78.67} & \textbf{100.00} & \textbf{79.88} & 30,797 & 70.40 \\
\midrule

\multirow{6}{*}{GPT-5.2} 
& MetaGPT      & 66.61 & \textbf{48.65} & 81.08 & 63.27 & 83.78 & 68.68 & 84,331 & 115.65 \\
& CAMEL        & 62.50 & 35.14 & 78.38 & 69.77 & 81.08 & 65.37 & 68,642 & \underline{87.46} \\
& CleanAgent    & 59.40 & 24.32 & 78.38 & 49.23 & 81.08 & 58.48 & \textbf{23,570} & \textbf{45.21} \\
& ChatDev2.0   & \underline{70.18} & 40.54 & \underline{83.78} & 74.42 & \underline{86.49} & 71.08 & 117,398 & 156.74 \\
& DataGovAgent     & 67.00 & \underline{43.24} & 81.08 & \underline{80.00} & \underline{86.49} & \underline{71.56} & 30,778 & 108.17 \\
& \textbf{ProfiliTable (ours)} & \textbf{82.49} & \textbf{48.65} & \textbf{97.30} & \textbf{96.55} & \textbf{100.00} & \textbf{85.00} & \underline{29,901} & 195.27 \\
\bottomrule
\end{tabular}
\end{table*}

In this subsection, we present a comprehensive performance comparison of ProfiliTable against strong baselines across single-step and multi-step table processing tasks. 

\textbf{Achieves SOTA in single-step tasks via dynamic profiling.}  
ProfiliTable attains state-of-the-art accuracy in single-step tasks across both base models. With GPT-4o, it achieves an ATS of 86.82, surpassing the second best method (MetaGPT) by 30.61 points. This large margin reflects its ability to unify data exploration, knowledge retrieval, and feedback-driven refinement through a dynamic profiling context. When paired with GPT-5.2, ProfiliTable further elevates ATS to 89.66, again far ahead of ChatDev2.0 (70.02). Crucially, this consistent gain is model-agnostic because the dynamic profiling context ensures robust reasoning regardless of base model capacity.

\textbf{Sets new SOTA in multi-step table processing.}  
ProfiliTable establishes state-of-the-art performance in multi-step tasks, achieving 80.19 (GPT-4o) and 82.49 (GPT-5.2) in ATS, while all baselines remain below 70.2. Critically, it is the only method to attain 100\% task-wise runnable rate (TRR) in both settings, which ensures that every task's final output is executable and meets a critical requirement for production deployment. The dynamic profiling context enables this capability by aligning subtask decomposition and execution with the global intent, thereby preventing error cascades.

\textbf{Highlights the fragility of baseline methods in the absence of unified profiling.} 
MetaGPT works on atomic tasks (ATS = 56.21 with GPT-4o) but collapses in multi-step workflows (ATS = 49.53), lacking compositional awareness. CAMEL performs poorly with GPT-4o but improves noticeably with GPT-5.2, revealing its heav\-y reliance on base model strength rather than robust task reasoning. In contrast, ProfiliTable performs consistently well on both models, demonstrating that its dynamic profiling context provides consistent, model-agnostic grounding for reliable table processing.

\textbf{Delivers high quality with low cost.}  
It uses only 24,907 tokens on average (GPT-5.2, single-step), which is the lowest among top performers, while maintaining acceptable runtime efficiency. By grounding retrieval and refinement within the profiling context, ProfiliTable avoids redundant actions and hallucinated code, demonstrating that dynamic profiling enables both accuracy and cost efficiency.

\subsection{Cost Analysis}
\begin{figure}[t]
    \centering
    \includegraphics[width=0.75\linewidth]{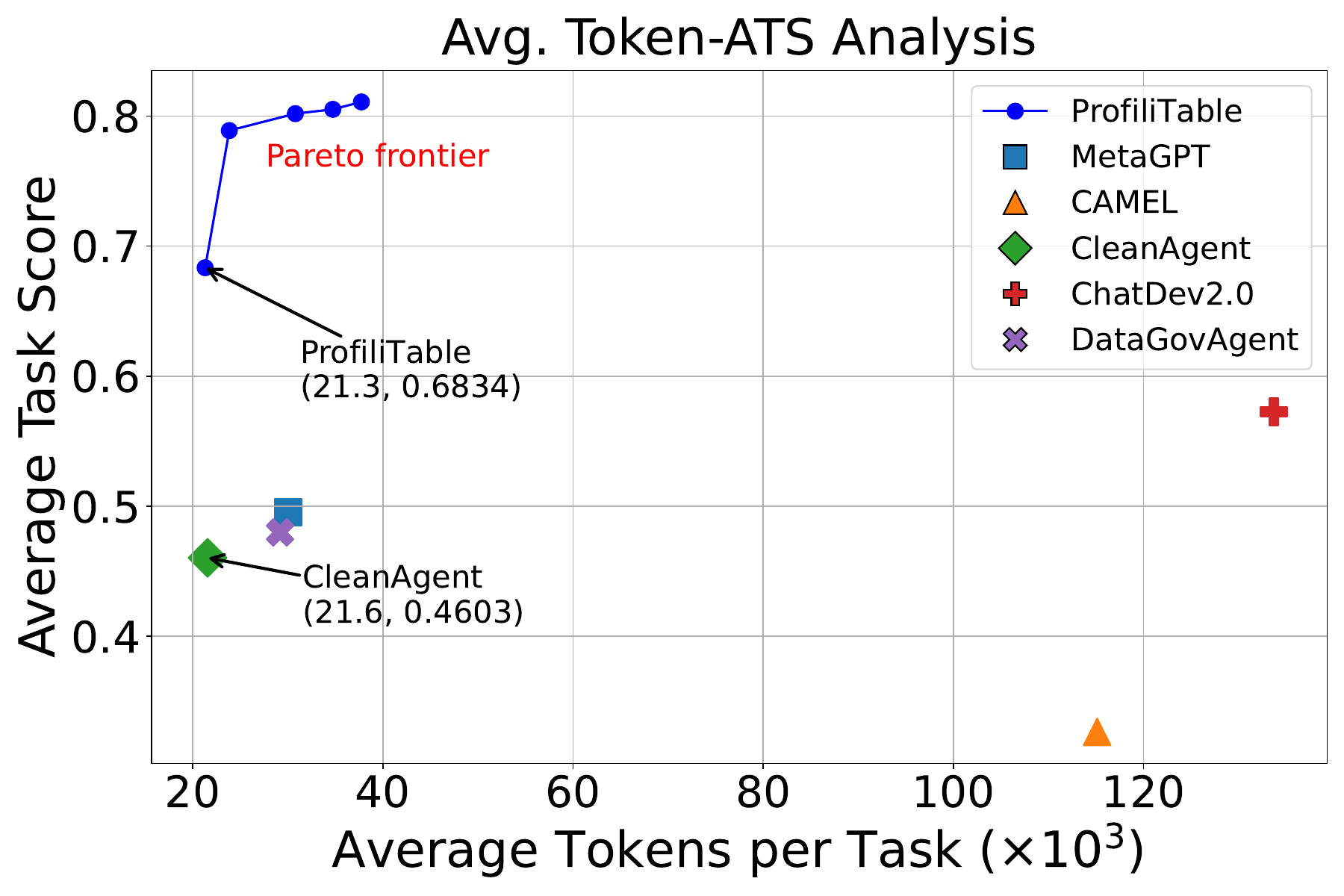}
    \caption{Trade-off between average token consumption and task accuracy (ATS) across methods on multi-step tasks with GPT-4o.}
    \label{fig:cost_analysis}
\end{figure}
\textbf{Traces a Pareto-optimal ATS–token trade-off.}  
We further analyze the efficiency–accuracy trade-off of ProfiliTable in terms of token cost and solution quality. As shown in Figure~\ref{fig:cost_analysis}, ProfiliTable traces an empirical Pareto frontier in the average token consumption versus ATS plane by varying the maximum refinement rounds \(T_{\max}\) from 1 to 5, with all other components fixed. Starting from a lightweight configuration (21.3k tokens, ATS = 68.34), it progressively improves performance as it invests additional tokens in structured decomposition, ReAct style profiling, and iterative refinement, reaching up to 81.09 ATS at 37.7k tokens. Crucially, all points along this trajectory form the Pareto frontier: CleanAgent achieves low token usage (21.6k)  but substantially lower accuracy (46.03); MetaGPT and DataGovAgent consume more tokens yet achieve only marginal gains in ATS; and ChatDev2.0 achieves higher ATS but incurs significantly higher costs. This demonstrates the strength of ProfiliTable: it delivers high-fidelity table transformations while maintaining strong cost efficiency, consistently occupying the optimal region of the performance--cost trade-off space.

\subsection{Hyperparameter Analysis}
\begin{figure}[t]
    \centering
    \begin{minipage}[b]{0.49\linewidth}
        \centering
        \includegraphics[width=\linewidth]{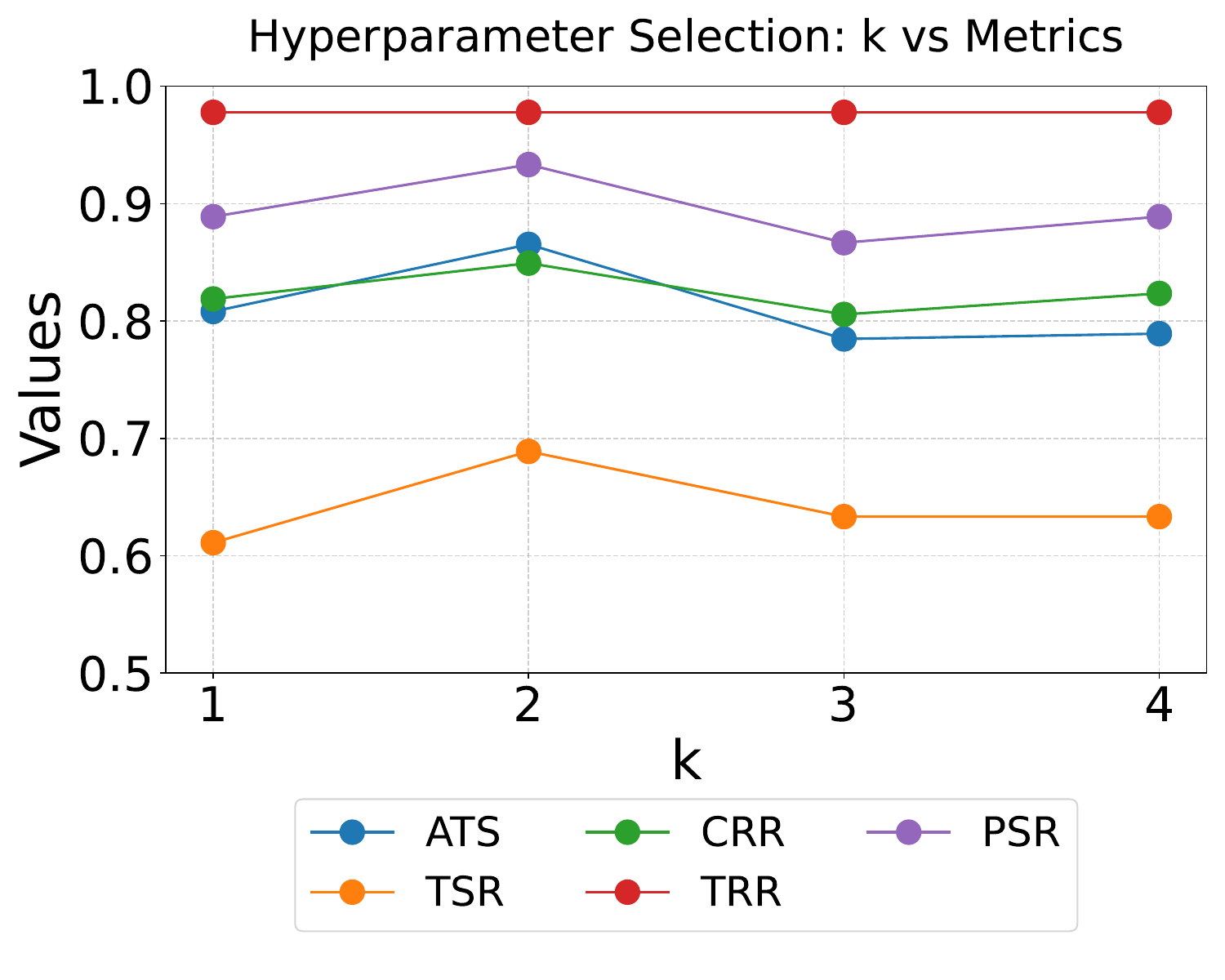}
        \subcaption{Hyperparameter: \(k\)}
        \label{fig:hyper_k}
    \end{minipage}
    \hfill
    \begin{minipage}[b]{0.49\linewidth}
        \centering
        \includegraphics[width=\linewidth]{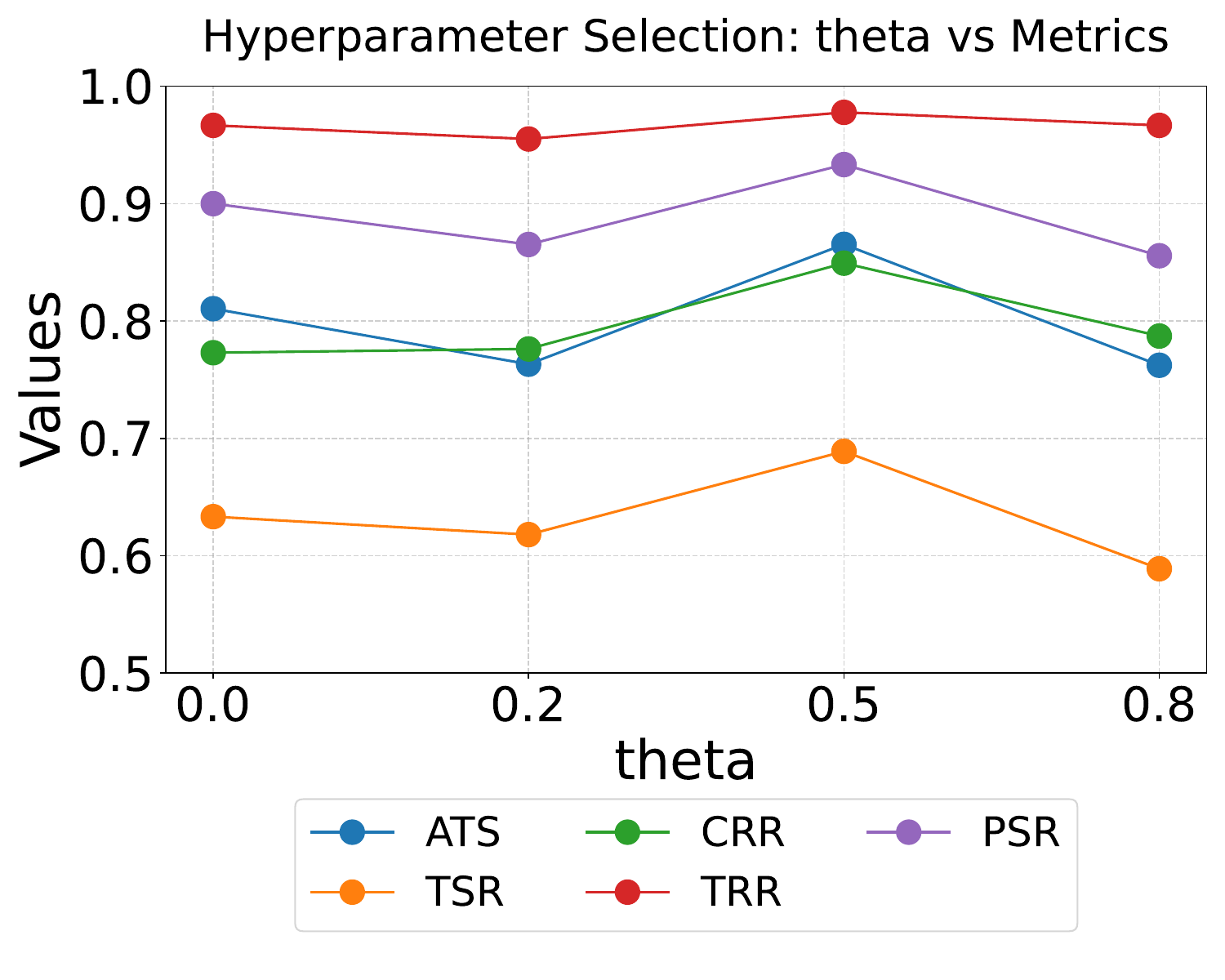}
        \subcaption{Hyperparameter: \(\theta_{\text{sim}}\)}
        \label{fig:hyper_theta}
    \end{minipage}
    \caption{Effect of hyperparameters $k$ and $\theta_{\text{sim}}$ on single-step performance (GPT-4o).}
    \label{fig:hyperparameters}
\end{figure}
This subsection analyzes the impact of two key RAG-related hyperparameters on ProfiliTable’s performance: the maximum number of retrieved operators \(k\) and the similarity threshold \(\theta_{\text{sim}}\) used during retrieval. Specifically, \(k\) controls the upper bound on how many operator templates are fetched from the knowledge base for a given task, while \(\theta_{\text{sim}}\) filters out low-relevance candidates by requiring a minimum semantic similarity between the task description and operator specification. Figure \ref{fig:hyperparameters} plots the trends of performance metrics as \(k\) and \(\theta_{\text{sim}}\) vary, respectively. The results reveal a clear trade-off between coverage and precision in retrieval, which we analyze in two aspects below.

\textbf{Retrieving at most \(k = 2\) operators achieves the best balance.} 
As shown in Figure \ref{fig:hyperparameters}, all metrics peak when \(k = 2\). Setting \(k = 1\) restricts the generator’s access to alternative implementations, reducing flexibility in handling ambiguous or multi-faceted tasks. In contrast, larger values (\(k = 3\) or \(4\)) introduce irrelevant or redundant operators, which increase prompt noise and raise the risk of hallucinated compositions, thereby degrading both correctness and robustness.

\textbf{A similarity threshold of \(\theta_{\text{sim}} = 0.5\) optimally filters signal from noise.} 
In Figure \ref{fig:hyperparameters}, performance is maximized at \(\theta_{\text{sim}} = 0.5\). Lower thresholds (e.g., \(\theta_{\text{sim}} = 0.2\)) admit too many semantically weak matches, overwhelming the generator with low-quality exemplars and leading to incorrect operator usage. Higher thresholds (e.g., \(\theta_{\text{sim}} = 0.8\)) are overly restrictive, often retrieving no operators for nuanced tasks—forcing the generator to fall back on generic code patterns that lack domain-specific grounding.

In conclusion, the configuration \(k = 2\) and \(\theta_{\text{sim}} = 0.5\) consistently yields the highest performance across all performance metrics, indicating that this setting provides sufficient diversity without sacrificing relevance. This validates our design principle: for table processing, effective RAG requires \textbf{focused retrieval}, which retrieves just enough relevant, high-quality operator examples to guide robust code generation without overwhelming the model with noisy or redundant candidates.

\subsection{Ablation Study}

\begin{figure}[t]
\centering
\begin{minipage}[b]{0.49\linewidth}
    \centering
    \includegraphics[width=\linewidth]{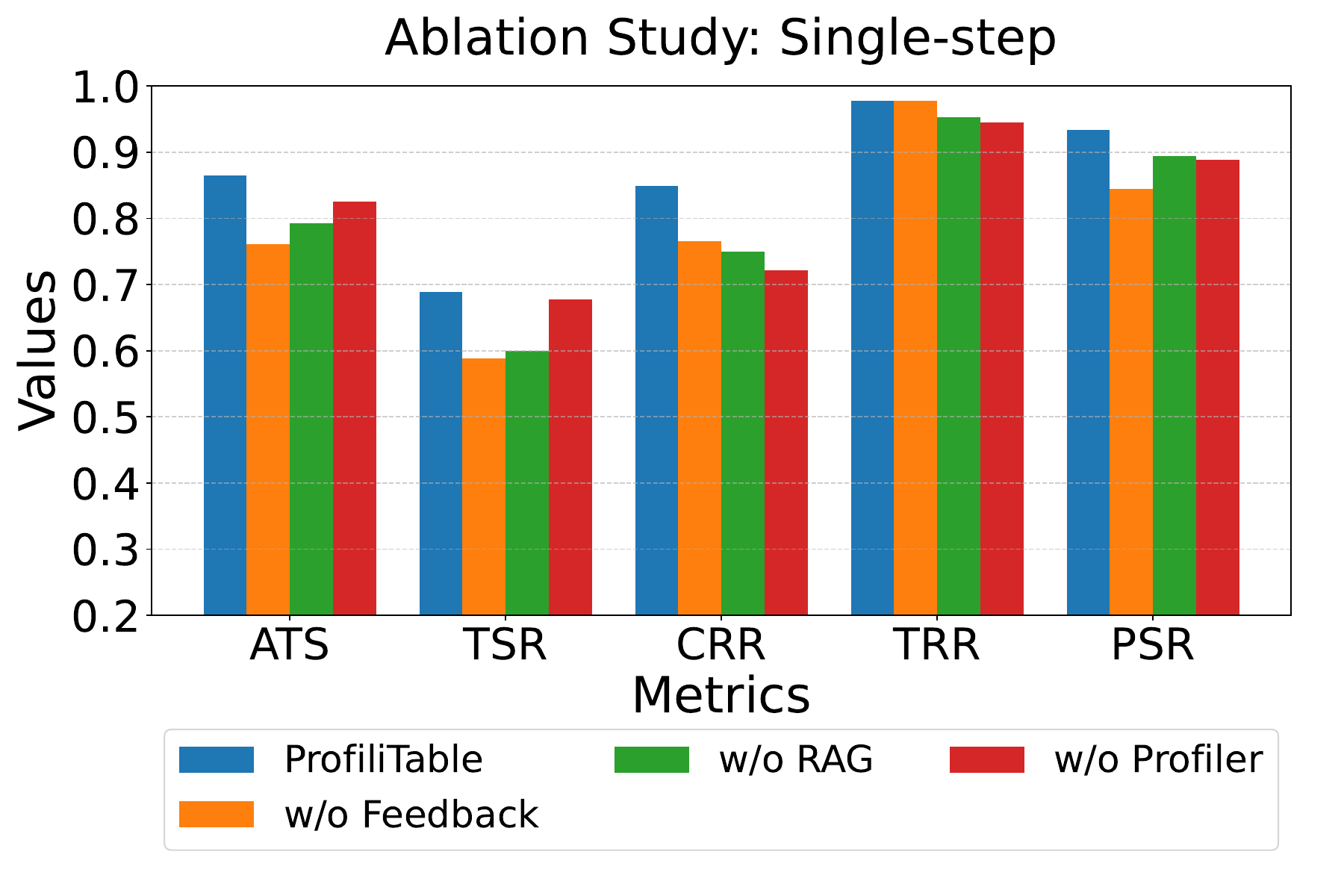}
    \subcaption{Single-step tasks}
    \label{fig:single-step}
\end{minipage}
\hfill
\begin{minipage}[b]{0.49\linewidth}
    \centering
    \includegraphics[width=\linewidth]{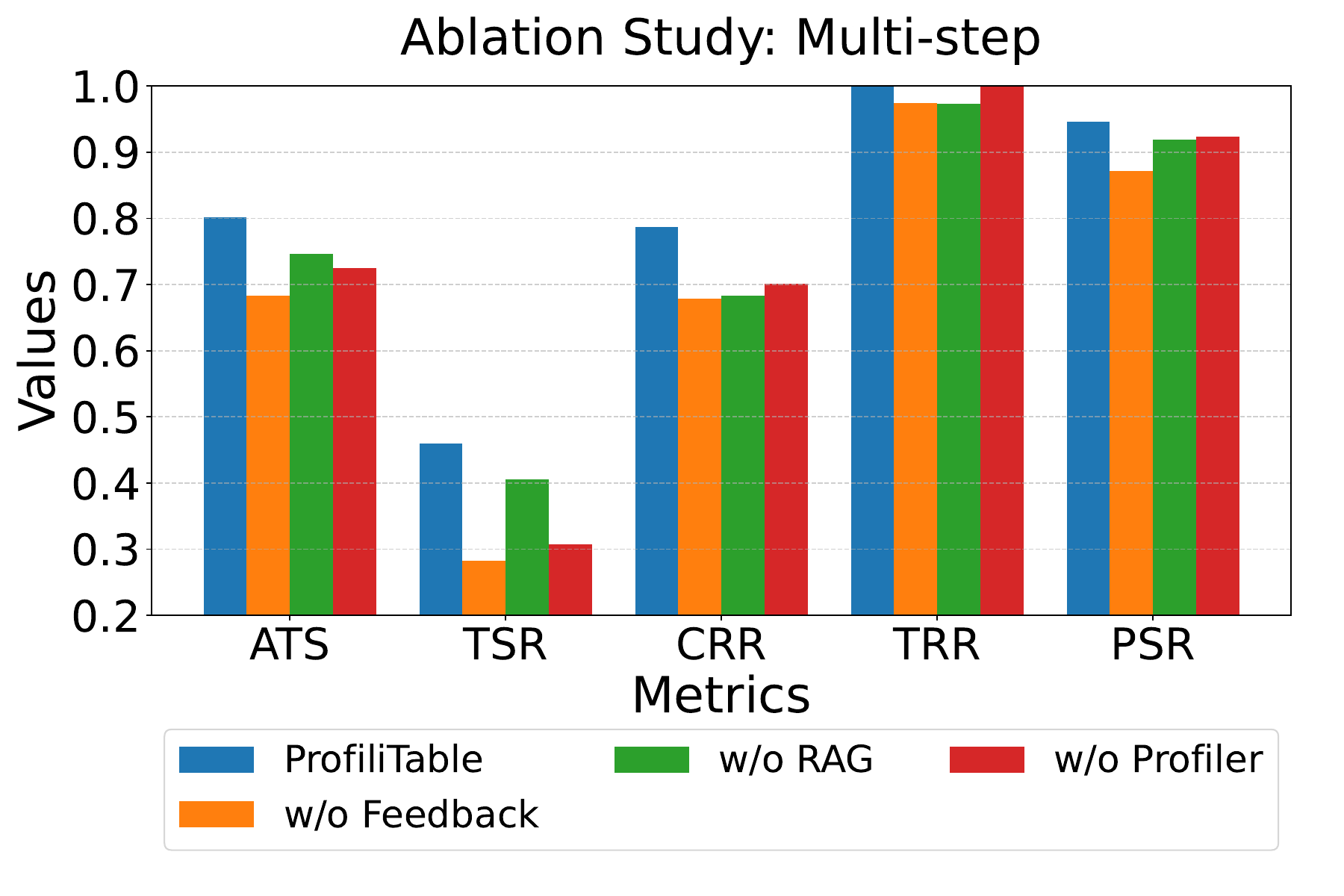}
    \subcaption{Multi-step tasks}
    \label{fig:multi-step}
\end{minipage}
\caption{Ablation study of ProfiliTable components across task complexities. Removing any module degrades performance, with the most severe drop observed when disabling feedback. The gap between full and ablated variants widens in multi-step settings (GPT-4o).}
\label{fig:ablation}
\end{figure}

\begin{figure}[t] 
  \centering
  \includegraphics[width=0.8\linewidth]{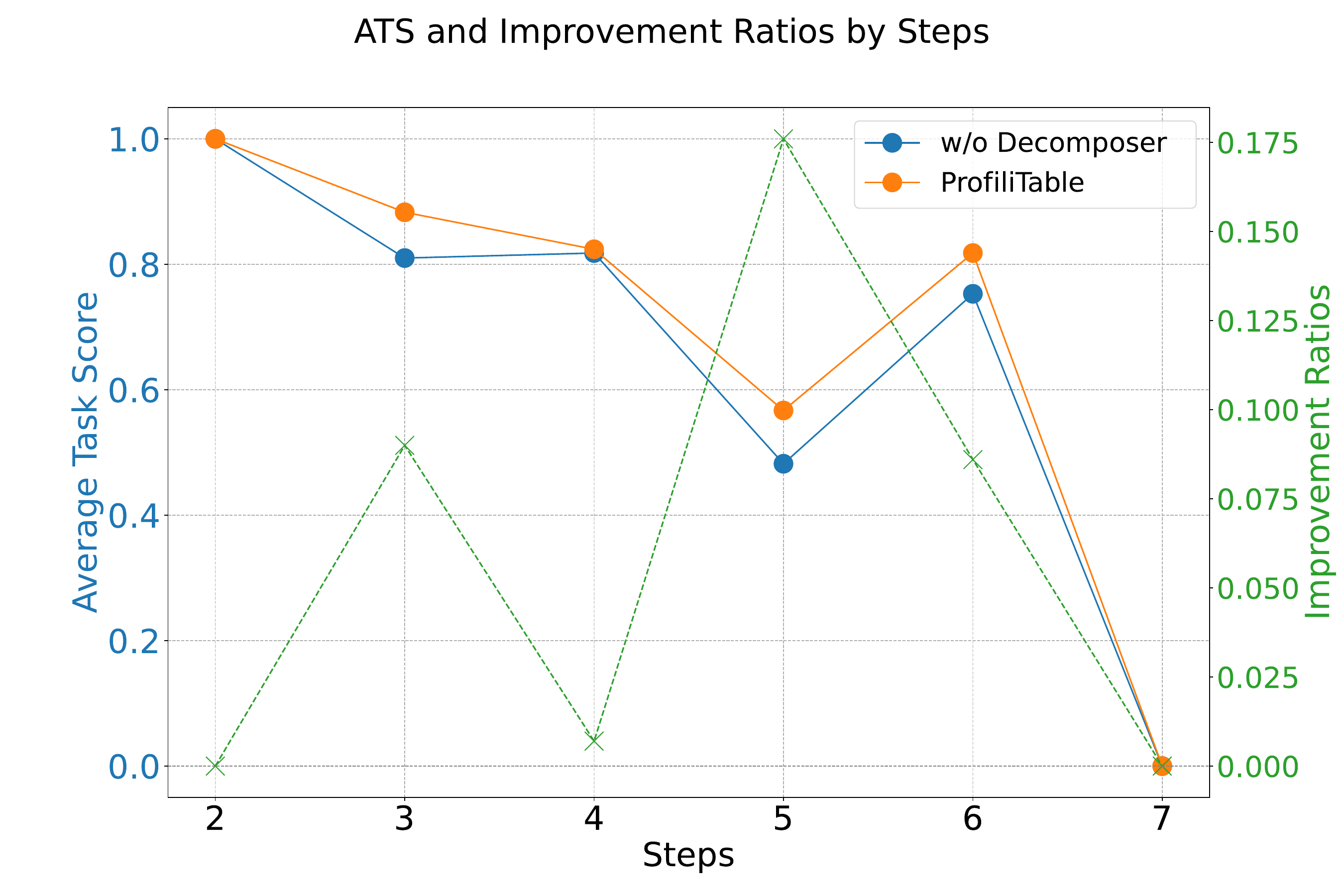}
  \caption{Impact of enabling or disabling the Decomposer module on performance across individual steps of multi-step tasks (GPT-4o).}
  \Description{Impact of enabling or disabling the Decomposer module on performance across individual steps of multi-step tasks.}
  \label{fig:Decomposer}
\end{figure}
To better understand how each component contributes to ProfiliTable’s strong performance, we conduct ablation studies on single-step and multi-step tasks, evaluating the impact of removing key modules: ReAct Profiler, Multi-round Feedback, RAG-enhanced Operator Retrieval, and the Decomposer for multi-step orchestration. These variants are denoted as \textit{w/o Profiler}, \textit{w/o Feedback}, \textit{w/o RAG}, and \textit{w/o Decomposer}, respectively. Results are summarized in Figure~\ref{fig:single-step} and Figure~\ref{fig:multi-step}, with a step-wise breakdown of the Decomposer’s effect shown in Figure~\ref{fig:Decomposer}.

\textbf{Profiling Module.} Removing the Profiler leads to a noticeable drop in all metrics, especially on multi-step tasks. In this case, the workflow uses a default, rule-based profiling strategy similar to that of DataGovAgent, which only collects basic column information without active exploration. This shows that static profiling cannot align context with task-specific requirements.

\textbf{Feedback Mechanism.}  
The \textit{w/o Feedback} variant, in which the Summarizer is disabled and the generator operates without iterative refinement, incurs the most severe performance degradation across all settings, particularly in ATS and TSR. On multi-step tasks, this gap is especially pronounced because the absence of reflective error diagnosis leads to uncorrected cascading failures that propagate through the workflow. Even on single-step tasks, the lack of post-execution insight results in consistent underperformance, underscoring feedback as the cornerstone of robust code generation.

\textbf{RAG Integration.}  
Disabling RAG (\textit{w/o RAG}) also reduces performance, particularly in ATS and CRR. This confirms that retrieving pre-validated operator templates helps suppress hallucination and improve correctness. However, the impact of removing RAG is comparatively milder on multi-step tasks than removing Profiling or Feedback. The performance drop is still noticeable, but smaller than that caused by ablating either the Profiler or the Summarizer. This suggests that, in complex workflows, structured exploration and iterative correction play a more decisive role than operator retrieval alone. Nevertheless, RAG remains valuable for grounding synthesis in reliable, domain-specific primitives.

\textbf{Decomposer for Task Decomposition.}  
To further dissect the role of structured decomposition, we perform a fine-grained ablation by disabling the Decomposer in multi-step tasks and measuring performance at each step (indexed 2–7). As shown in Figure~\ref{fig:Decomposer}, the improvement from enabling the Decomposer varies significantly across steps: it yields the largest improvements at intermediate levels of task complexity. For example at step 5, ATS increases from ~0.482 to ~0.567 (+17.5\% relative improvement), indicating that explicit subtask breakdown is most beneficial when precise operator sequencing is required. In contrast, early steps (2–3) see minimal gains, as atomic operations can often be handled directly by the generator. At step 7, both variants collapse to zero performance due to accumulated errors, revealing a current limitation in ultra-deep pipelines. This step-wise analysis confirms that the Decomposer acts as a \textbf{targeted enhancer}—its value emerges precisely when task structure becomes nontrivial but not yet overwhelmed by error propagation.

Critically, the full workflow consistently outperforms all ablated variants, with performance gaps widening significantly on multi-step tasks. This demonstrates that \textbf{dynamic profiling} becomes increasingly essential as task complexity grows. These results validate our design: by addressing diverse failure modes in table processing, dynamic profiling enables reliable and scalable automation.

\subsection{Limitations and Implications}
Reliable table processing demands more than strong models: it requires structured, reflective workflows centered on \textbf{dynamic profiling}. Real-world tables often suffer from schema ambiguity, causing non-profiling methods to fail even with advanced LLMs. 

\textbf{Failure Analysis.} We identify 3 primary failure modes: \textit{Code or Structure Errors} (syntax or shape mismatches), \textit{Content Errors} (value discrepancies), and the \textit{Compliance Trap} (blindly following noisy instructions). Our analysis shows that while scaling mitigates Code Errors, high-capability models remain constrained by Content and Structure errors, indicating a lack of micro-level logical precision. Furthermore, the ``Compliance Trap'' reveals that models often prioritize literal intent over data integrity when faced with ambiguous inputs. 

\textbf{Implications.} These limitations underscore that generic LLMs lack the micro-level logical precision required for rigorous table manipulation. To bridge this gap, future work should focus on table-specific fine-tuning, adapting models to internalize table conventions and structural constraints, thereby enhancing their fidelity and robustness in table-centric tasks.

\section{Conclusion}
We present \textbf{ProfiliTable}, a multi-agent framework for autonomous table processing built around \textbf{dynamic profiling}. Evaluated with GPT-4o and GPT-5.2, it outperforms all baselines by wide margins and is the only method to achieve 100\% task-wise runnable rate in multi-step settings, ensuring that every task's output is executable—a critical requirement for deployment. It also lies on the Pareto frontier of accuracy and efficiency, showing that dynamic profiling delivers both high fidelity and low cost. Our work points to a new paradigm: \textbf{profiling-driven agency}, which refers to iterative, interactive, and failure-aware systems that treat tables as dynamic, semantically rich artifacts.


\begin{acks}
  This work is supported by the National Natural Science Foundation of China (U23B2048, U22B2037). Bin Cui and Wentao Zhang are the corresponding authors.
\end{acks}

\bibliographystyle{ACM-Reference-Format}
\bibliography{sample-base}


\appendix

\section{Additional Discussions}
This section presents the results of DeepAnalyze (Section~\ref{subsec: add-exp}) and discusses future directions (Section~\ref{subsec: future-work}).

\subsection{Additional Experiment} \label{subsec: add-exp}
In this subsection, we present additional results for DeepAnalyze \cite{DeepAnalyze}. We include this method in the appendix rather than the main evaluation because it represents a fundamentally different paradigm: DeepAnalyze is built around a single, domain-adapted language model—specifically, ds-0528-qwen3-8b fine-tuned on chain-of-thought (CoT) data for data science tasks—rather than a multi-agent framework with distinct, collaborating roles. Its core mechanism extends the ReAct framework by introducing a set of special tokens that condition the model’s behavior at each step. Depending on the observed state and user intent, the model selects actions, such as coding and analyzing, by emitting these tokens, effectively implementing an enhanced, token-driven ReAct loop. While this design enables flexible data analysis, it lacks the explicit division of labor, interactive feedback, and profiling-driven coordination that characterize true multi-agent systems like ProfiliTable. 

We further extend our evaluation to include additional strong backbones (DeepSeek-V3, Claude Opus 4.6, and Qwen3-Coder-\linebreak[4]Flash) and SheetMind, a framework specialized for spreadsheet operations included here due to its narrow scope. As shown in Table \ref{tab:extended-single}, Table \ref{tab:extended-multi}, Table~\ref{tab:deepanalyze}, ProfiliTable consistently outperforms all these methods across diverse settings, maintaining its leading performance while incurring only moderate computational costs. To further address generalizability concerns, we evaluated ProfiliTable on the DataGovBench benchmark. ProfiliTable leads with ATS of (89.23, 90.40, 90.82, 82.90, 76.18) across the five models (GPT-4o, GPT-5.2, Claude Opus 4.6, DeepSeek-V3, Qwen3-Coder-Flash), significantly surpassing strong baselines including CleanAgent (46.47, 52.21, 59.54, 53.70, 53.56), MetaGPT (62.99, 70.35, 62.11, 36.88, 51.02), ChatDev2.0 (60.60, 74.49, 67.53, 54.43, 54.38), CAMEL (51.10, 62.34, 51.39, 50.79, 56.16), DataGovAgent (53.98, 69.13, 67.13, 43.37, 36.49), and SheetMind (40.17, 52.44, 51.45, 52.86, 53.48). These results confirm that ProfiliTable’s dynamic profiling paradigm delivers robust performance not only on our comprehensive benchmark but also on established external benchmarks across diverse model backbones.

\subsection{Future Work} \label{subsec: future-work}
An exciting direction is to leverage the high-quality CoT \cite{CoT} data synthesized by ProfiliTable’s workflow to enhance single-model capabilities. Specifically, we plan to use reinforcement learning algorithms such as GRPO \cite{GRPO} to fine-tune base language models on this curated dataset. This approach would yield table-processing-specific models that inherit both the structured reasoning of our multi-agent framework and the efficiency of a unified architecture, combining the best of workflow-level intelligence and model-level specialization.


\section{Metrics} \label{subsec:metric}
As shown in Table~\ref{tab:metrics}, we provide a detailed description of the evaluation metrics, where $N$ denotes the total number of evaluated tasks. These metrics collectively assess both the \textit{effectiveness} and \textit{efficiency} of table processing agents: performance-oriented measures (e.g., ATS, TSR, PSR, CRR, TRR) capture correctness, robustness, and partial progress, while efficiency-oriented measures (e.g., Avg. Tokens and Avg. Time) quantify computational cost and latency, offering a holistic view of agent capabilities in real-world table wrangling scenarios.

\begin{table*}[ht]
\centering
\small
\caption{Performance comparison on single-step tasks. Best results are in \textbf{bold}, and second-best results are \underline{underlined}.}
\label{tab:extended-single}
\begin{tabular}{lccccccccc}
\toprule
\textbf{Base Model} & \textbf{Framework} & \textbf{ATS$\uparrow$} & \textbf{TSR$\uparrow$} & \textbf{PSR$\uparrow$} & \textbf{CRR$\uparrow$} & \textbf{TRR$\uparrow$} & \textbf{Avg. Score$\uparrow$} & \textbf{Avg. Tokens$\downarrow$} & \textbf{Avg. Time (s)$\downarrow$} \\
\midrule

\multirow{7}{*}{Claude Opus 4.6} 
& CleanAgent      & 57.11 & 50.00 & 60.00 & 54.69 & 68.89 & 58.14 & \underline{21,171} & \textbf{19.67} \\
& MetaGPT         & 62.11 & 51.61 & 64.52 & 60.40 & 67.74 & 61.28 & 103,167 & 122.84 \\
& ChatDev2.0      & \underline{65.73} & \underline{52.22} & \underline{71.11} & 73.30 & \underline{81.11} & 68.69 & 34,671 & 131.69 \\
& CAMEL           & 51.39 & 41.94 & 54.84 & 52.52 & 58.06 & 51.75 & 42,125 & 166.13 \\
& DataGovAgent    & 62.70 & 51.11 & 66.67 & \textbf{90.12} & 75.56 & \underline{69.23} & 57,097 & 168.60 \\
& SheetMind       & 46.59 & 40.00 & 50.00 & 56.56 & 58.89 & 50.41 & \textbf{9,409} & \underline{66.47} \\
& \textbf{ProfiliTable (ours)} & \textbf{88.44} & \textbf{77.78} & \textbf{91.11} & \underline{87.60} & \textbf{97.78} & \textbf{88.54} & 58,914 & 162.26 \\
\midrule

\multirow{7}{*}{Deepseek-V3} 
& CleanAgent      & 50.20 & 40.00 & 54.44 & 45.99 & 64.44 & 51.01 & 17,832 & \textbf{21.41} \\
& MetaGPT         & 34.67 & 28.89 & 37.78 & 47.37 & 50.00 & 39.74 & 87,667 & 170.93 \\
& ChatDev2.0      & \underline{55.93} & \underline{41.11} & \underline{61.11} & 53.04 & 76.67 & \underline{57.57} & 74,883 & 169.80 \\
& CAMEL           & 46.15 & 37.78 & 52.22 & 53.60 & 65.56 & 51.06 & \underline{15,475} & \underline{40.66} \\
& DataGovAgent    & 33.89 & 23.33 & 47.78 & \underline{64.54} & \underline{80.00} & 49.91 & 33,328 & 125.16 \\
& SheetMind       & 47.21 & 37.78 & 51.11 & 49.70 & 70.00 & 51.16 & \textbf{4,128} & 41.01 \\
& \textbf{ProfiliTable (ours)} & \textbf{80.32} & \textbf{65.56} & \textbf{85.56} & \textbf{72.04} & \textbf{85.56} & \textbf{77.81} & 27,812 & 159.85 \\
\midrule

\multirow{7}{*}{Qwen3-Coder-Flash} 
& CleanAgent      & 50.17 & 38.89 & 56.67 & 48.46 & 76.67 & 54.17 & 20,149 & \textbf{10.08} \\
& MetaGPT         & 46.37 & 34.44 & 53.33 & 48.46 & 70.00 & 50.52 & 193,930 & 120.19 \\
& ChatDev2.0      & \underline{56.22} & \underline{43.33} & \underline{64.44} & 58.62 & 75.56 & 59.63 & 127,980 & 235.45 \\
& CAMEL           & 53.98 & 38.89 & \underline{64.44} & 63.39 & \underline{78.89} & \underline{59.92} & 223,846 & 106.60 \\
& DataGovAgent    & 33.66 & 26.67 & 37.78 & 65.49 & 66.67 & 46.05 & 30,797 & 53.58 \\
& SheetMind       & 49.72 & 37.78 & 55.56 & \textbf{70.54} & 70.00 & 56.72 & \textbf{3,860} & \underline{17.46} \\
& \textbf{ProfiliTable (ours)} & \textbf{75.08} & \textbf{60.67} & \textbf{83.15} & \underline{70.11} & \textbf{91.01} & \textbf{76.00} & \underline{16,499} & 36.27 \\
\bottomrule
\end{tabular}
\end{table*}

\begin{table*}[ht]
\centering
\small
\caption{Performance comparison on multi-step tasks. Best results are in \textbf{bold}, and second-best results are \underline{underlined}.}
\label{tab:extended-multi}
\begin{tabular}{lccccccccc}
\toprule
\textbf{Base Model} & \textbf{Framework} & \textbf{ATS$\uparrow$} & \textbf{TSR$\uparrow$} & \textbf{PSR$\uparrow$} & \textbf{CRR$\uparrow$} & \textbf{TRR$\uparrow$} & \textbf{Avg. Score$\uparrow$} & \textbf{Avg. Tokens$\downarrow$} & \textbf{Avg. Time (s)$\downarrow$} \\
\midrule

\multirow{7}{*}{Claude Opus 4.6} 
& CleanAgent      & 63.80 & 37.84 & 72.97 & 69.82 & 72.97 & 63.48 & \underline{22,240} & \textbf{21.21} \\
& MetaGPT         & \underline{77.83} & 45.95 & \underline{91.89} & 72.02 & \underline{91.89} & \underline{75.92} & 90,234 & 131.77 \\
& ChatDev2.0      & 74.35 & 43.24 & 83.78 & 77.90 & 83.78 & 72.61 & 37,205 & 143.43 \\
& CAMEL           & 47.56 & 27.03 & 51.35 & 72.41 & 51.35 & 49.94 & 39,558 & 203.53 \\
& DataGovAgent    & 65.62 & \underline{46.15} & 76.92 & \underline{89.39} & 84.62 & 72.54 & 63,761 & 224.32 \\
& SheetMind       & 69.26 & \textbf{48.65} & 78.38 & 78.95 & 81.08 & 71.26 & \textbf{15,963} & \underline{76.60} \\
& \textbf{ProfiliTable (ours)} & \textbf{83.64} & \textbf{48.65} & \textbf{97.30} & \textbf{100.00} & \textbf{93.85} & \textbf{84.69} & 99,800 & 301.37 \\
\midrule

\multirow{7}{*}{Deepseek-V3} 
& CleanAgent      & 57.62 & 27.03 & 64.86 & 52.63 & 64.86 & 53.40 & \underline{18,314} & \textbf{21.12} \\
& MetaGPT         & 53.29 & \underline{32.43} & 64.86 & 54.69 & 70.27 & 55.11 & 75,195 & 149.14 \\
& ChatDev2.0      & \underline{66.79} & \textbf{37.84} & \underline{81.08} & \underline{68.42} & \underline{86.49} & \underline{68.12} & 77,971 & 206.51 \\
& CAMEL           & 33.67 & 18.92 & 40.54 & 67.86 & 54.05 & 43.01 & 23,255 & 64.07 \\
& DataGovAgent    & 38.55 & 20.51 & 51.28 & 60.00 & 58.97 & 45.86 & 35,423 & 215.83 \\
& SheetMind       & 62.71 & 29.73 & 75.68 & 44.44 & 81.08 & 49.93 & \textbf{4,690} & \underline{40.92} \\
& \textbf{ProfiliTable (ours)} & \textbf{78.78} & 29.73 & \textbf{94.59} & \textbf{76.62} & \textbf{97.30} & \textbf{75.40} & 31,124 & 184.77 \\
\midrule

\multirow{7}{*}{Qwen3-Coder-Flash} 
& CleanAgent      & 47.43 & 29.73 & 54.05 & 48.24 & 54.05 & 46.70 & 20,888 & \underline{24.21} \\
& MetaGPT         & 52.39 & 29.73 & 64.86 & 57.14 & 75.68 & 55.96 & 207,213 & 114.96 \\
& ChatDev2.0      & \underline{64.32} & \underline{37.84} & 81.08 & 69.41 & \underline{83.78} & 67.29 & 117,229 & 232.56 \\
& CAMEL           & 59.95 & 24.32 & \textbf{83.78} & 74.07 & \textbf{94.59} & \underline{67.34} & 244,971 & 119.71 \\
& DataGovAgent    & 31.58 & 15.38 & 43.59 & \underline{77.19} & 48.72 & 43.29 & 36,583 & 69.14 \\
& SheetMind       & 47.20 & 18.92 & 64.86 & 73.81 & 75.68 & 56.09 & \textbf{4,503} & \textbf{22.49} \\
& \textbf{ProfiliTable (ours)} & \textbf{72.52} & \textbf{38.89} & \underline{83.33} & \textbf{80.00} & 83.33 & \textbf{71.61} & \underline{19,650} & 44.36 \\
\bottomrule
\end{tabular}
\end{table*}

\begin{table*}[ht]
\centering
\caption{Performance of DeepAnalyze on single-step and multi-step tasks}
\label{tab:deepanalyze}
\begin{tabular}{l l c c c c c c}
\toprule
Framework & Task Type & ATS$\uparrow$ & TSR$\uparrow$ & PSR$\uparrow$ & TRR$\uparrow$  & Avg. Time (s)$\downarrow$\\
\midrule
DeepAnalyze & single-step & 44.47 & 32.22 & 54.44 & 85.56 &  39.36 \\
\midrule
DeepAnalyze & multi-step  & 54.35 & 16.22 & 72.97 & 91.89 & 65.35 \\
\bottomrule
\end{tabular}
\end{table*}

\begin{table*}[ht]
\centering
\caption{Evaluation metrics used in this work}
\label{tab:metrics}
\begin{tabular}{l l l p{6.5cm}}
\toprule
\textbf{Metric} & \textbf{Abbr.} & \textbf{Formula} & \textbf{Description} \\
\midrule
\multicolumn{4}{l}{\textit{Performance Metrics (reported as percentages)}} \\
\midrule
Average Task Score & ATS & $100 \times \frac{1}{N}\sum_{i=1}^{N} \text{score}_i$ & Mean of normalized task scores (0–1), scaled to percentage; measures overall correctness. \\
Task Success Rate & TSR & $100 \times \frac{1}{N}\sum_{i=1}^{N} \mathbb{I}(\text{score}_i = 1)$ & Percentage of tasks achieving perfect output (full score of 1.0). \\
Partial Success Rate & PSR & $100 \times \frac{1}{N}\sum_{i=1}^{N} \mathbb{I}(\text{score}_i > 0)$ & Percentage of tasks with strictly positive scores, indicating meaningful partial progress. \\
Code Runnable Rate & CRR & $100 \times \frac{\#\text{runnable scripts}}{\#\text{generated scripts}}$ & Percentage of generated code snippets that execute successfully (syntactic and runtime validity). \\
Task-Wise Runnable Rate & TRR & $100 \times \frac{\#\text{tasks with at least one runnable script}}{N}$ & Percentage of tasks for which at least one attempt yields a runnable script. \\
Average Metric Score & Avg. Score & $ \frac{\text{ATS} + \text{TSR} + \text{PSR} + \text{CRR} + \text{TRR}}{5}$ & Mean of the five core performance metrics (ATS, TSR, PSR, CRR, TRR), providing a holistic assessment of correctness, robustness, and reliability. \\

\midrule
\multicolumn{4}{l}{\textit{Efficiency Metrics}} \\
\midrule
Average Tokens & Avg. Tokens & $\frac{1}{N}\sum_{i=1}^{N} \text{tokens}_i$ & Average number of tokens consumed by the agent per task. \\
Average Time & Avg. Time (s) & $\frac{1}{N}\sum_{i=1}^{N} t_i$ & Average wall-clock execution time per task in seconds. \\
\bottomrule
\end{tabular}
\end{table*}

\section{Prompt Design}
\label{subsec:prompts}

This appendix presents the system prompt templates used by the major
modules in ProfiliTable. Each module is assigned a role-specific prompt
that constrains its objective, reasoning behavior, and output format.

\begin{PromptBox}[promptGray]{System Prompt for Interpreter}
\footnotesize
\begin{alltt}
[ROLE]
You are a Table preprocessing intent parsing API. Based on the 
metadata of the data table \{task_meta\}, parse the user's natural 
language instruction into a standardized JSON format to identify the 
required data processing operator. You must respond with only a JSON 
object—do not wrap it in ```json.

[OUTPUT RULES]:
1. **operation**: Clearly describe the required operator operation in 
natural language, use orders like 1. ..., 2. ... to list multiple 
operations if needed. You should describe as detailed as possible.
2. **reason**: Briefly explain the rationale for selecting this 
operator (1–2 sentences), possibly referencing missing rates, 
data distribution, or task objectives in the metadata.
3. **is_dag**: Specify whether the task involves exactly one operator 
task (see task types below) or multiple operator tasks. If multiple, 
the operations should be executed in a Directed Acyclic Graph (DAG) 
manner rather than sequentially. Output true for DAG and false for 
sequential.
4. **task_type**: If the task involves a single operator, select 
exactly one matching task type from: 

"TableCleaning-ErrorDetectionANDCorrection",
"TableCleaning-ColumnTypeAnnotation",
"TableCleaning-DataImputation",
"TableCleaning-Deduplication",

# Table Transformation
"TableTransformation-RowToRowTransform",
"TableTransformation-SplittingANDConcatenation", 
"TableTransformation-RowColumnSwapping",
"TableTransformation-Filtering",
"TableTransformation-Grouping",
"TableTransformation-Sorting",
"TableTransformation-ListExtraction",

# Table Augmentation
"TableAugmentation-RowPopulation",
"TableAugmentation-SchemaAugmentation",
"TableAugmentation-ColumnAugmentation",

# Table Matching
"TableMatching-SchemaMatching",
"TableMatching-EntityMatching" 

The chosen type must strictly align with both the table metadata and 
the user's intent.
5. **suffix**: Specify the output file format, e.g., "csv", "jsonl", 
etc.

Your output must be a valid JSON object only, with no additional text, 
explanations, Markdown formatting (such as ```json), or line breaks. 
Strictly follow the format in the example below.

[Example]:
User request: Fill missing values in the age column using an LSTM 
model trained on other columns
Your output: 
\{"operation": "1:fill missing values in the column age using LSTM  
trained on other columns", "reason": "'age' has 30\% missing values, 
which may impact analysis. Using LSTM can better capture relationship 
with other columns to impute missing values.", "task_type": 
"TableCleaning-DataImputation", "suffix": "csv", "is_dag": false\}

User request: sorting and filtering tables based on some conditions
Your output: 
\{"operation": "1:sorting the table based on some conditions, 
2:filtering the table based on some conditions", "reason": "The user 
request involves two distinct operations that can be executed 
sequentially. Sorting can help organize the data, while filtering can 
refine the results based on specific criteria.", "suffix": "csv", 
"is_dag": true, "task_type": "TableTransformation-Sorting"\}
\end{alltt}
\end{PromptBox}

\begin{PromptBox}[promptGray]{System Prompt for Decomposer}
\footnotesize
\begin{alltt}
[ROLE]
You are an expert in decomposing complex tasks into independent,
executable sub-tasks.

[OUTPUT RULES]
1. Decompose the task into independent sub-tasks.
2. Each sub-task should be executable and must be drawn from
\{benchmark_task_types\}.
3. Output the result strictly in JSON format, mapping each sub-task
type to its specific operation description, with no ```json wrapping.
4. Each key in the JSON should be a different sub-task type, and the
corresponding value should be the specific operation description.

[EXAMPLE]
User request: Merge multiple CSV files and deduplicate entries based 
on a primary key.
Your output:
\{"TableTransformation-SplittingANDConcatenation": "Merge multiple 
CSV files", "TableCleaning-Deduplication": "Deduplicate entries 
based on a primary key"\}

Note: ensure the output keys correspond to different sub-task types.
Concatenate multiple operations under the same sub-task type into one
description if needed.
\end{alltt}
\end{PromptBox}

\begin{PromptBox}[promptGray]{System Prompt for Profiler}
\footnotesize
\begin{alltt}
[ROLE]
You are a careful data profiler to prepare a data report for the 
target. Your role is to write code that analyzes tabular data files  
and produces a comprehensive data profiling report in JSON format. 
You are given up to \{MAX_REACT_STEPS\} attempts to reach a 
conclusion.

[INPUTS]
File paths: \{raw_table_paths\}
Target: \{operation\}

[GOAL]
Prepare for what the target requires by analyzing the data files and
producing a detailed profiling report. You are not to fulfill the
target yet; only analyze and report for further processing. 
Produce the final report as JSON inside
<ANSWER>\{"table_1": \{...\}, "table_2": \{...\}, ...\}</ANSWER>,
where table_x is replaced by the filename without extension
(e.g., sales.csv -> "sales"). At minimum, include the number of rows,
number of columns, column names, column types, and abnormal types 
like mixed types or unexpected nulls. If the goal targets a particular
column or transformation, analyze that column in detail, including
unique values, missing rate, and numeric distribution if relevant. If
the number of unique values or some other statistics is small (e.g.,
less than 20), list them all; otherwise, sample 5--10 values. The
final report should be concise (do not surpass 200 characters unless
necessary) but comprehensive, focusing on key statistics and useful
insights while avoiding irrelevant information.

[RULES]
In each turn:
- Use <THINK>...</THINK> to describe your reasoning.
- Use <ACTION>```python ... ```</ACTION> to provide standalone,
  executable Python code.
- After the action, wait for the observation (the printed output).
- After receiving observation, continue reasoning with <THINK>, then
  issue the next <ACTION> if needed.
- Your code must be fully self-contained, including all imports, data
  loading, and logic.
- Always load data from the provided file paths.
- Use print() to output results and avoid printing raw data or huge
  outputs.
- For multiple files, profile each one and include a separate entry in
  the final JSON report.
- Keep code precise and concise (<=50 lines per action unless
  absolutely necessary).
- Do not write any files to disk; only output via print().
- Once profiling is complete, output the full report in
  <ANSWER>...</ANSWER> as valid JSON without ``` formatting.

[EXAMPLE]
<THINK>I need to read the first CSV file and get basic column info.
</THINK>
<ACTION>
```python
import pandas as pd
df = pd.read_csv("data/sales.csv")
print(\{"columns": list(df.columns), "shape": list(df.shape)\})
```
</ACTION>
[Observation]
\{"columns": ["id", "amount", "date"], "shape": [1000, 3]\}
<THINK>Now I'll compute statistics for numeric columns...</THINK>
<ACTION>
```python
import pandas as pd
df = pd.read_csv("data/sales.csv")
numeric_cols = df.select_dtypes(include='number').columns
stats = df[numeric_cols].describe().to_dict()
print(stats)
```
</ACTION>
<THINK>All tables are profiled. Compiling final JSON report.</THINK>
<ANSWER>\{...\}</ANSWER>
\end{alltt}
\end{PromptBox}

\begin{PromptBox}[promptGray]{System Prompt for Generator}
\footnotesize
\begin{alltt}
[ROLE]
You specialize in table processing. Please generate a bug-free Python
script based on the following information and the user's request.

[INPUT]
1. Metadata of the data table: \{task_meta\}
2. Retrieved similar operator code snippets: \{retrieved_operators\}
3. Operator specification: \{user_query\}

[OUTPUT RULES]
1. The code must be executable, safe, step by step, and output as a
complete code block in the format ```python ... ```.
2. The code must include a main() function that accepts command-line
arguments for input and output file paths. Use a fixed argparse format
with two required arguments: --input (input file paths)
and --output_path_dir (output file path directory).
3. The function must fulfill all user requirements. Ensure that the
output file format matches the user's request and contains no extra
columns beyond those in the input.
4. If the task involves multiple tables, the --input argument should 
be treated as a list of file paths. The --output_path_dir argument 
should be the directory where the results will be saved.
5. Avoid modifying the original input files; read from them and write
results to new files in the specified output directory.
6. No BOM should be used in output CSV files; do not save with
encoding='utf-8-sig'.
7. Keep the code step by step rather than compressing complex logic
into a single block. Use as many steps as needed to ensure clarity and
correctness.

[EXAMPLE]
[INPUT]
User request:
Fill missing values in the age column using an LSTM model trained on
other columns
Operator specification:
\{"operation": "fill missing values in the column age using an LSTM
model trained on other columns", "reason": "'age' has 30\% missing
values, which may impact analysis. Using an LSTM can better capture
relationships with other columns to impute missing values.",
"task_type": "TableCleaning-DataImputation", "suffix": "csv"\}
Metadata:
\{...\}
Retrieved similar operator code snippets:
[ ... ]
Debug history:
[ ... ]

[OUTPUT] (illustrative only)
```python
import argparse
import pandas as pd

def fill_missing_age_with_lstm(df):
    return df

def main():
    parser = argparse.ArgumentParser()
    parser.add_argument("--input", required=True, nargs='+')
    parser.add_argument("--output_path_dir", required=True)
    args = parser.parse_args()
    ...

if __name__ == "__main__":
    main()
```
\end{alltt}
\end{PromptBox}

\begin{PromptBox}[promptGray]{System Prompt for Debugger}
\footnotesize
\begin{alltt}
[ROLE]
You are an expert in code debugging and correction.

[TASK]
Given the original code, error message, requirement, and reference
code, minimally modify the original code to fix the error. Ensure that
your corrections are precise and focus on issues such as key alignment
or import errors. Output the corrected code and your reason for
modification strictly in JSON format and follow all specified
requirements.

[INPUT]
You will receive the following information in the human request:
- The original code
- The error messages
- The target
- Raw data and expected data formats

[OUTPUT RULES]
1. The response must be strictly in JSON format, containing only the
keys "code" (with the complete corrected code) and "reason"
(explaining the modification); no extra keys, explanations, comments
, or Markdown syntax are allowed.
2. The code's --input and --output_path_dir arguments should be kept
unchanged.
3. The code must include an if __name__ == '__main__': block to ensure
the script can be run independently.
4. All parser arguments should have default values except --input and
--output_path_dir.
5. Your output must be a valid JSON object only, with no additional
text, explanations, Markdown formatting (such as ```json), or line
breaks.
6. No additional files or external references should be included 
unless explicitly required to resolve the error, and if needed, 
they must be handled within the code itself.
\end{alltt}
\end{PromptBox}

\begin{PromptBox}[promptGray]{System Prompt for Summarizer}
\footnotesize
\begin{alltt}
[ROLE]
You are an evaluator tasked with analyzing the processed results
of a task to determine whether they meet the target requirements. Your
role is to write code that evaluates the processed file(s) and gives
a summary of whether the target requirements are satisfied, and gives
reasonable suggestions for improvement if not. You are given up to
\{MAX_REACT_STEPS\} attempts to reach a conclusion.

[INPUTS]
Metadata: \{task_meta\}
Processed file paths: \{processed_file_paths\}
Raw file paths: \{raw_file_paths\}
Task objective: \{task_objective\}

[GOAL]
Generated code should directly assess the content of the processed
file(s) for basic reasonableness, such as the presence of required
fields, structural or schema consistency, and the absence of obvious
anomalies (e.g., empty arrays, malformed JSON or CSV, and unexpected
nulls in critical columns). Do not generate ground truth or simulate
expected outputs. Do not compute, assign, or justify any numerical
scores. Do not attempt to modify the files, nor check whether a
hypothetical fix worked. Identify concrete, observable issues and, if
present, give short, actionable advice. Sample each column's data
from the processed files and compare it with the raw files to identify
discrepancies relevant to the task objective. If inspection reveals 
no clear issues, or further analysis yields conclusions nearly 
identical to the previous round, promptly summarize the result and 
output
<ANSWER>.

[RULES]
In each turn:
- Use <THINK>...</THINK> to describe your reasoning.
- Use <ACTION>```python ... ```</ACTION> to provide standalone,
  executable Python code.
- Use <ANSWER>...</ANSWER> to provide the final evaluation summary 
as a string.
- After the action, wait for the observation (the printed output).
- After receiving observation, continue reasoning with <THINK>, then
  issue the next <ACTION> if needed.
- Your code must be fully self-contained, including all imports, data
  loading, and logic.
- Always load data from the provided file paths.
- Use print() to output results and avoid huge outputs.
- Keep code precise and concise (<=50 lines per action unless
  absolutely necessary).
- Do not write any files to disk; only output via print().
- Once you find problems or determine that all requirements are met,
  output the final summary in <ANSWER>...</ANSWER> immediately.

[EXAMPLE]
<THINK>I need to load the processed file and check whether it meets 
the target requirements.</THINK>
<ACTION>
```python
import pandas as pd
df = pd.read_csv("processed_file.csv")
if "target_column" in df.columns:
    print("Evaluation result: Pass")
else:
    print("Evaluation result: Fail, missing 'target_column'")
```
</ACTION>
[Observation]
Evaluation result: Fail, missing 'target_column'
<THINK>The processed file does not meet the target requirements. I 
will check the remaining conditions.</THINK>
...
<ANSWER>The processed file misses 'target_column' and ...</ANSWER>
\end{alltt}
\end{PromptBox}

\end{document}